\def\year{2022}\relax
\documentclass[letterpaper]{article} 
\usepackage{aaai22}  
\usepackage{times}  
\usepackage{helvet} 
\usepackage{courier}  
\usepackage[hyphens]{url}  
\usepackage{graphicx} 
\usepackage{natbib}  
\usepackage{caption} 
\usepackage{amsmath}
\usepackage{mathbbol}
\usepackage{nicefrac}       
\usepackage[all,cmtip]{xy}  
\usepackage{graphbox} 
\usepackage[labelformat=simple]{subcaption}

\usepackage{bm}
\usepackage[noabbrev]{cleveref}

\DeclareMathOperator{\im}{im}

\DeclareMathOperator{\Ltp}{\tilde{\mathcal{L}}^+} 
\DeclareMathOperator{\hL}{\mathcal{L}} 
\newcommand{\Ltpd}[1]{\tilde{\mathcal{L}}^+_{#1}} 
\newcommand{\cmplx}{0.215}
\newcommand{\cmplxx}{0.11}
\newcommand{\errorplt}{0.248}
\newcommand{\scaleplt}{0.23}
\newcommand{\cmplxD}{0.215}
\newcommand{\cmplxDD}{0.12}

\title{Dist2Cycle: A Simplicial Neural Network for Homology Localization}
\author{Alexandros D. Keros\textsuperscript{\rm 1}, Vidit Nanda\textsuperscript{\rm 2}, Kartic Subr\textsuperscript{\rm 1}\\}
\affiliations{
    \textsuperscript{\rm 1}The University of Edinburgh\\
    \textsuperscript{\rm 2}University of Oxford\\
    a.d.keros@sms.ed.ac.uk, nanda@maths.ox.ac.uk, ksubr@ed.ac.uk
}

\begin{document}

\maketitle

\begin{abstract}
Simplicial complexes can be viewed as high dimensional generalizations of graphs that explicitly encode multi-way ordered relations between vertices at different resolutions, all at once. This concept is central towards detection of higher dimensional topological features of data, features to which graphs, encoding only pairwise relationships, remain oblivious. While attempts have been made to extend Graph Neural Networks (GNNs) to a simplicial complex setting, the methods do not inherently exploit, or reason about, the underlying topological structure of the network. We propose a graph convolutional model for learning functions parametrized by the $k$-homological features of simplicial complexes. By spectrally manipulating their combinatorial $k$-dimensional Hodge Laplacians, the proposed model enables learning topological features of the underlying simplicial complexes, specifically, the distance of each $k$-simplex from the nearest ``optimal" $k$-th homology generator, effectively providing an alternative to homology localization.

\end{abstract}


\begin{figure*}[htbp]
    \centering
    \includegraphics[width=\textwidth]{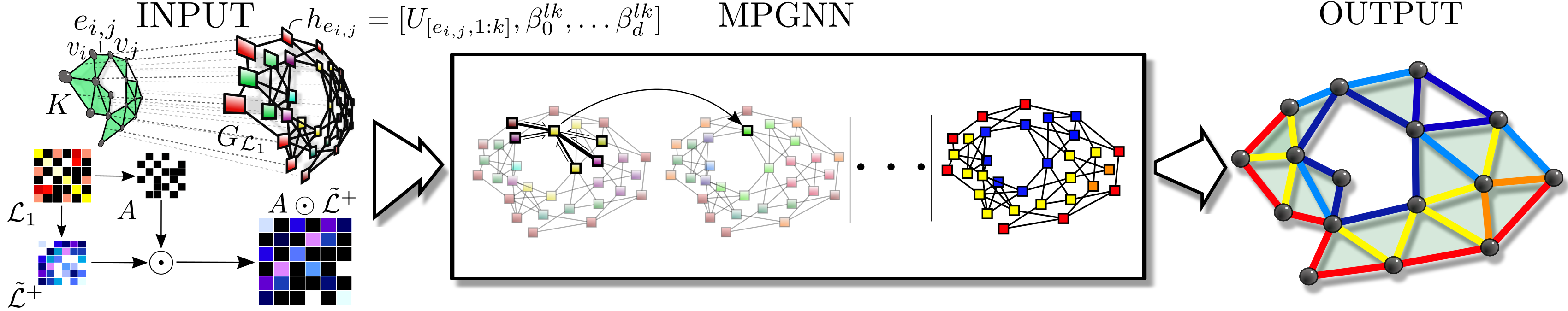}
    \caption{Overview of Dist2Cycle model. An arbitrary input complex $K$ is transformed into a Hodge Laplacian graph $G_{\mathcal{L}_1}$, via its Hodge Laplacian appropriately shift-inverted ($\tilde{\mathcal{L}}^+$) (Section~\ref{subsec:Ltp}), and zero-masked ($A\odot \tilde{\mathcal{L}}^+$) (Section~\ref{subsec:SCGNN}), suitable for downstream processing as a GNN by the proposed Dist2Cycle model (Section~\ref{subsec:SFGNN}). The model outputs the distance of each simplex to its nearest optimal homology generator (cooler colors indicate closeness).}
    \label{fig:model_overview}
\end{figure*}

\section{Introduction}\label{sec:Intro}
Tremendous advancements in sensor technology and data-driven machine learning have enabled exciting applications such as automatic health monitoring and autonomous cars. In many cases, the lack of data in certain regions of the domain reveals important structure. For instance, the sensors on a car driving through a parking lot might have dense observation points in 3D except inside pillars. Such \emph{voids} in the data are ubiquitous across applications whether it is a subspace of unattainable configurations for a robot~\cite{farber2018configuration}, regions without network coverage~\cite{Ghrist2005} or missing measurements in an experimentally-determined chemical structure~\cite{townsend2020representation}, to name a few~\cite{aktas2019persistence}.  

A standard approach to extract structural information from data proceeds by first encoding pairwise relationships in a problem via a graph and then analysing its properties.  Recent advances in Graph Neural Networks have enabled practical learning in the domain of graphs and have provided approximate solutions to difficult graph  problems~\cite{hamilton2017representation}. Despite the wealth of techniques underpinned by solid graph theory, this approach is fundamentally  misaligned with problems where the relationships involve multiple points, and topological \& geometric structure must be encoded beyond pairwise interactions.

Fortunately, higher dimensional combinatorial structures come to the rescue in the form of simplicial complexes, the powerhorse of topological data analysis~\cite{chazal2017introduction}. Interfacing between combinatorics and geometry, simplicial complexes capture multi-scale relationships and facilitate the passage from local structure to global invariant features. These features occur in the form of homology groups, intuitively perceived as {\em holes}, or {\em voids}, in any desired dimension. Alas, this expressive power comes with a burden, that of high computational complexity, and difficulty in localization of said voids~\cite{chen2011hardness}.

For every hard computational problem there seems to exist a neural network approximation~\cite{xu2018powerful}. Nevertheless, homology and simplicial complexes have only recently started to follow suit~\cite{bodnar2021weisfeilerMPNNS, ebli2020simplicial,bunch2020simplicial}, with inference {\em of} homological information still lacking.

The key insight in this paper is to guide learning on simplicial complexes by flipping the conventional view of approximation. We propose a GNN model for localizing homological information in the form of a distance function of each point of a complex to its the nearest homology generating features, a bird's-eye view of which is illustrated in Figure~\ref{fig:model_overview}. 
Instead of using the most-significant eigenvectors of the relevant Laplacian operator we focus on the subspace spanned by the eigenvectors corresponding to the lowest eigenvalues. The justification is that homology-related information is contained in its nullspace. We implement this idea by calculating the most-significant subspace of an inverted version of the operator (see Sec.~\ref{subsec:Ltp}). Figure~\ref{fig:diff} shows the result of twelve diffusion iterations performed using the conventional view and compares it with our inverted operator. Although diffusion is insufficient to localise homology, it highlights the tendency of the inverted operator to localize cycles. 

The main contributions in the paper are:
\begin{enumerate}
    \item A novel way to represent simplicial complexes as computational graphs suitable for inference via GNNs, the {\em Hodge Laplacian graphs}, focusing on the dimension of interest (see Section~\ref{subsec:SCGNN}), and
    \item A new homology-aware graph convolution framework operating on the proposed Hodge Laplacian graphs, taking advantage of the spectral properties of a shifted-inverted version of the Hodge Laplacian (see Section~\ref{subsec:SFGNN}, Eq.~\eqref{eq:SFGC}).
\end{enumerate}

The rest of the paper is structured as follows: first all the necessary theoretical background is presented in the ``Preliminaries" section, followed by a literature review of relevant work. We then describe our proposed model in detail in the ``Dist2Cycle" section. We end the paper with a thorough evaluation and discussion of our model.

\section{Preliminaries}\label{sec:Prelims}

\newcommand{\Ldsize}{0.11}
\begin{figure*}[t]
     \centering
     \begin{subfigure}[b]{\Ldsize\textwidth}
         \centering
         \includegraphics[width=\textwidth]{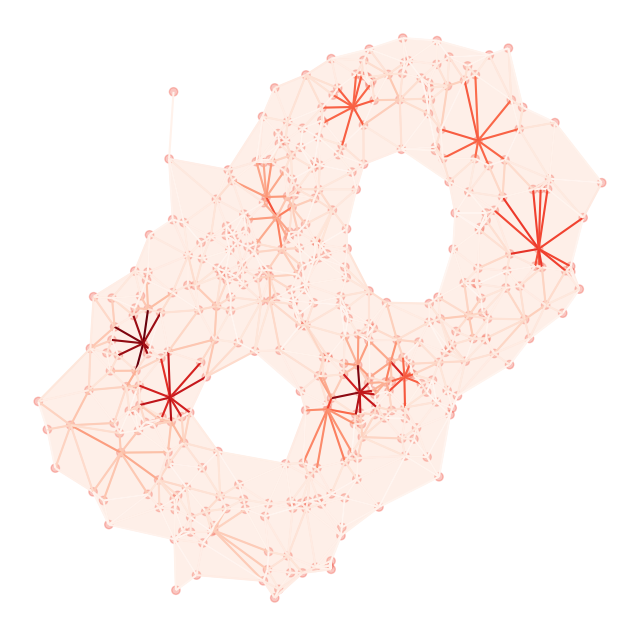}
         \caption{$\hL_1$}
         \label{fig:hL1}
     \end{subfigure}
     \hfill
     \begin{subfigure}[b]{\Ldsize\textwidth}
         \centering
         \includegraphics[width=\textwidth]{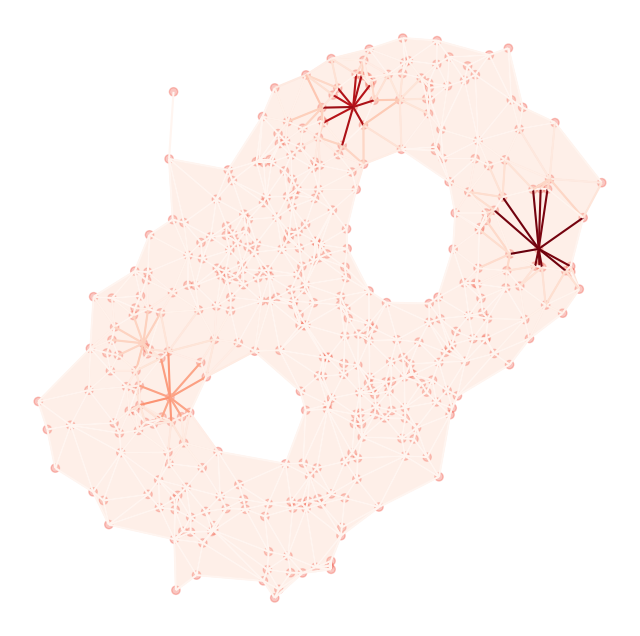}
         \caption{$\hL_1$ (rank-3)}
         \label{fig:hL1_k3}
     \end{subfigure}
     \hfill
     \begin{subfigure}[b]{\Ldsize\textwidth}
         \centering
         \includegraphics[width=\textwidth]{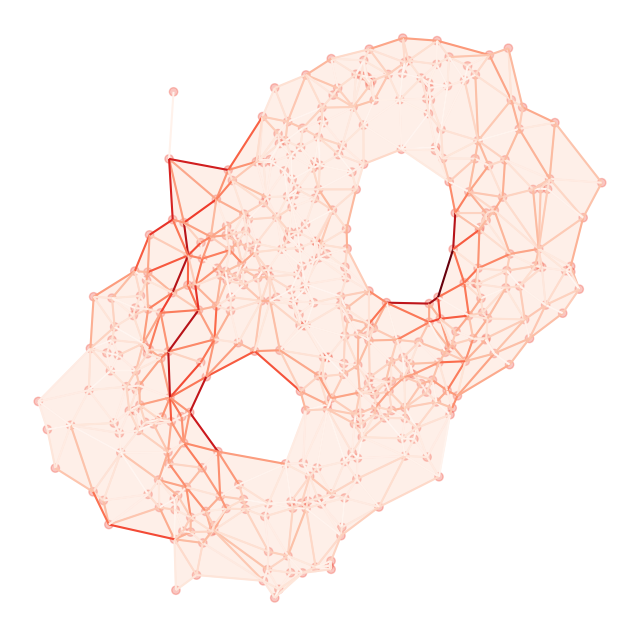}
         \caption{$\Ltpd{1}$}
         \label{fig:Ltpd1}
     \end{subfigure}
     \hfill
     \begin{subfigure}[b]{\Ldsize\textwidth}
         \centering
         \includegraphics[width=\textwidth]{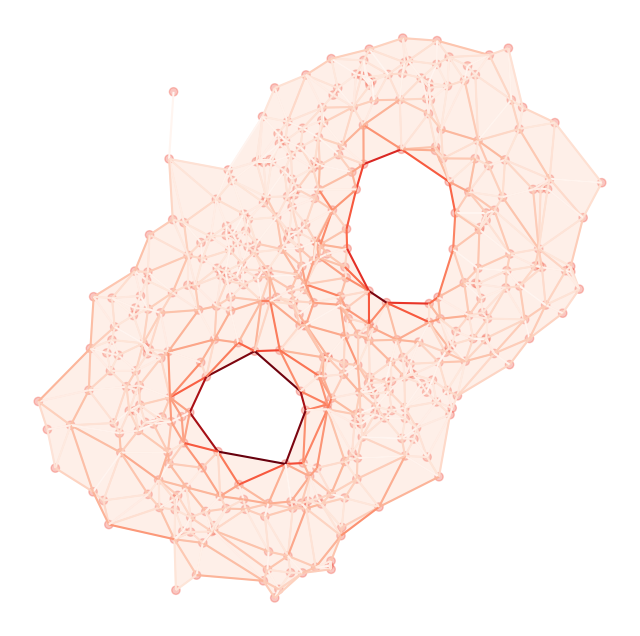}
         \caption{$\Ltpd{1}$(rank-3)}
         \label{fig:Ltpd1_k3}
     \end{subfigure}
    \hfill
     \begin{subfigure}[b]{\Ldsize\textwidth}
         \centering
         \includegraphics[width=\textwidth]{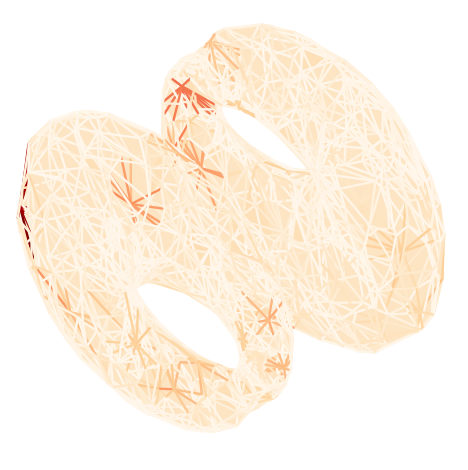}
         \caption{$\hL_1$}
         \label{fig:hL1_3D}
     \end{subfigure}
     \hfill
     \begin{subfigure}[b]{\Ldsize\textwidth}
         \centering
         \includegraphics[width=\textwidth]{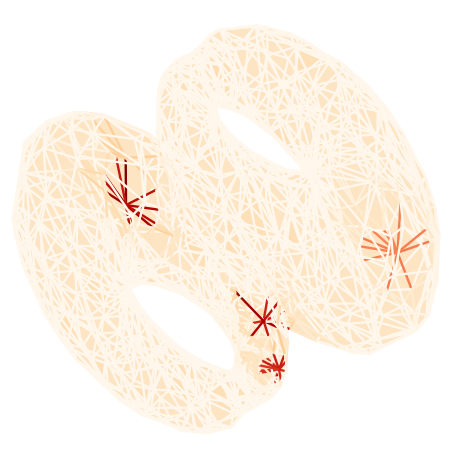}
         \caption{$\hL_1$ (rank-5)}
         \label{fig:hL1_3D_k5}
     \end{subfigure}
     \hfill
     \begin{subfigure}[b]{\Ldsize\textwidth}
         \centering
         \includegraphics[width=\textwidth]{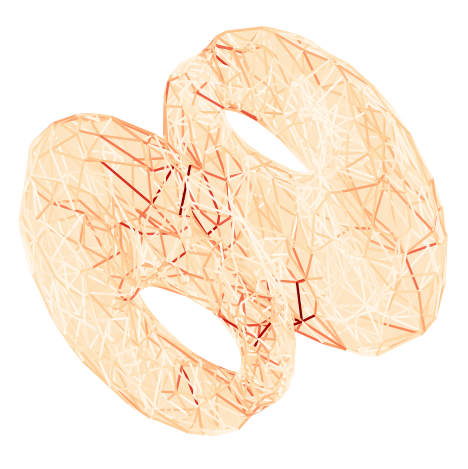}
         \caption{$\Ltpd{1}$}
         \label{fig:Ltpd1_3D}
     \end{subfigure}
     \hfill
     \begin{subfigure}[b]{\Ldsize\textwidth}
         \centering
         \includegraphics[width=\textwidth]{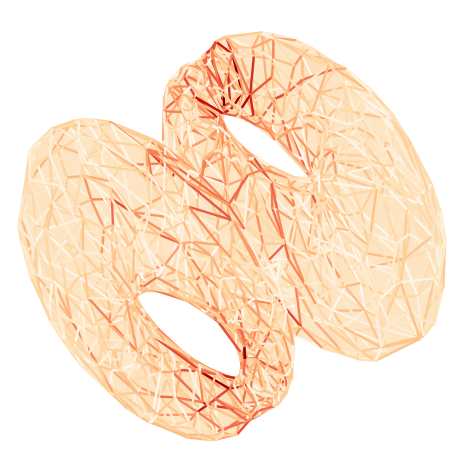}
         \caption{$\Ltpd{1}$(rank-5)}
         \label{fig:Ltpd1_3D_k5}
     \end{subfigure}
        \caption{\label{fig:diff}Twelve diffusion iterations based on the 1-dimensional Hodge Laplacian $\hL_{1}$, and its shifted pseudoinverse $\Ltpd{1}$ of a double annulus in 2D (four leftmost) and a double torus in 3D (four rightmost). \subref{fig:hL1_k3}, \subref{fig:Ltpd1_k3}, \subref{fig:hL1_3D_k5}, and \subref{fig:Ltpd1_3D_k5} are low-rank approximations of the respective Laplacians based on the top 3 and 5 eigenpairs corresponding to the largest magnitude eigenvalues. 
        }
        \label{fig:laplacian_diffusion}
\end{figure*}
\subsection{The Simplicial Laplacian Operators}\label{subsec:homology}

An {\em abstract simplicial complex} is a collection $K$ of subsets of a finite set $S$ satisfying two axioms: first, for each $v$ in $S$ the singleton set $\{v\}$ lies in $K$, and second, whenever some $\sigma \subset S$ lies in $K$, every subset of $\sigma$ must also lie in $K$. The constituent subsets $\sigma \subset S$ which lie in $K$ are called {\em simplices}, and the dimension of each such $\sigma$ is one less than its cardinality, i.e., $\dim \sigma = |\sigma|-1$. By far the most familiar examples of simplicial complexes are (undirected, simple) {\em graphs}; each graph $G = (V,E)$ forms a simplicial complex whose $0$-dimensional simplices are given by the vertex set $V$ and $1$-dimensional simplices constitute the edge set $E$. The passage from graphs to simplicial complexes is motivated by the compelling desire to model phenomena beyond pairwise interactions using higher-dimensional simplices.

\subsubsection{Homology Groups}

To each directed graph $G=(V,E)$ one can associate an {\em incidence matrix}, which is best viewed as a linear map $A:\mathbb{R}[E] \to \mathbb{R}[V]$ from a real vector space spanned by edges to the vector space spanned by the vertices. The entry of $A$ in the column corresponding to a directed edge $e:v \to v'$ and the row corresponding to a vertex $u$ is prescribed by
\[
A_{u,e} = \begin{cases} -1  & \text{if } u = v, \\
                        1 & \text{ if } u = v', \text{ and} \\
                        0 & \text{ otherwise.}
        \end{cases}.
\]
Writing $r$ for the rank of $A$, the number of connected components and loops in $G$ equals $|V|-r$ and $|E|-r$, respectively. Thus, one can learn the geometry of $G$ from the linear algebraic data given by its adjacency matrix.

This linear algebraic success story admits a remarkable simplicial sequel. Fix a simplicial complex $K$ and write $K_d$ to indicate the set of all $d$-simplices in $K$. We seek linear maps $\partial_d:\mathbb{R}[K_d] \to \mathbb{R}[K_{d-1}]$ to play the role of the $d$-dimensional incidence matrices. To build these {\em boundary operators}, one first orders the vertices in $K_0$ so that each $d$-simplex $\sigma \in K$ can be uniquely expressed as a list $\sigma = [v_0,\ldots,v_d]$ of vertices in increasing order. The desired matrix $\partial_d$ is completely prescribed by the following action on each such $\sigma$:
\begin{align}\label{eq:sbound}
\partial_d(\sigma) = \sum_{i=0}^d (-1)^i \cdot  \sigma_{-i}
\end{align}
where $\sigma_{-i} := [v_0,\dots,\hat{v}_i,\ldots,v_d]$ is the $(d-1)$-simplex obtained by removing the $i$-th vertex $v_i$ from $\sigma$.

These higher incidence operators assemble into a sequence of vector spaces and linear maps:
\begin{align}\label{eq:chcomp}
\xymatrix{
\cdots 
\ar@{->}[r]^--{\partial_{d+1}} 
& \mathbb{R}[K_{d}] \ar@{->}[r]^{\partial_{d}} & \mathbb{R}[K_{d-1}] \ar@{->}[r]^--{\partial_{d-1}} & \cdots.
}
\end{align}
It follows from \eqref{eq:sbound} that for each $d > 0$ the composite $\partial_d \circ \partial_{d+1}$ is the zero map, so the kernel of $\partial_d$ contains the image of $\partial_{d+1}$ as a subspace, $\im\partial_{d+1} \subseteq \ker \partial_{d}$.

For each $d \geq 0$, the $d$-th {\bf homology group} of $K$ is the quotient vector space $\mathcal{H}_d(K) := {\ker \partial_d}/{\im \partial_{d+1}}$ of $k$-cycles $\mathcal{Z}_k=\ker \partial_d$ by $(k+1)$-boundaries  $\mathcal{B}_k=\im \partial_{d+1}$. The basis of $\mathcal{H}_d(K)$ contains equivalence classes of $d$-dimensional {\em voids} or {\em loops} $[g_i]$, i.e. $\mathcal{H}_d(K)=\text{span}\{[g_1], \dots, [g_k]\}$, each $[g_i]$ describing a family of loops that cannot be contracted to a point, and cannot be continuously deformed into another family $[g_j]$, $i\neq j$.  Consequently, the dimension of $\mathcal{H}_d(K)$ provides us with a topological invariant, namely, the $k$-th \textbf{betti number} $\beta_d=\text{rank}(\mathcal{H}_d(K))$, which counts the number of $d$-dimensional voids in $K$. 

\begin{figure}[b]
    \centering
    \includegraphics[width=0.2\textwidth]{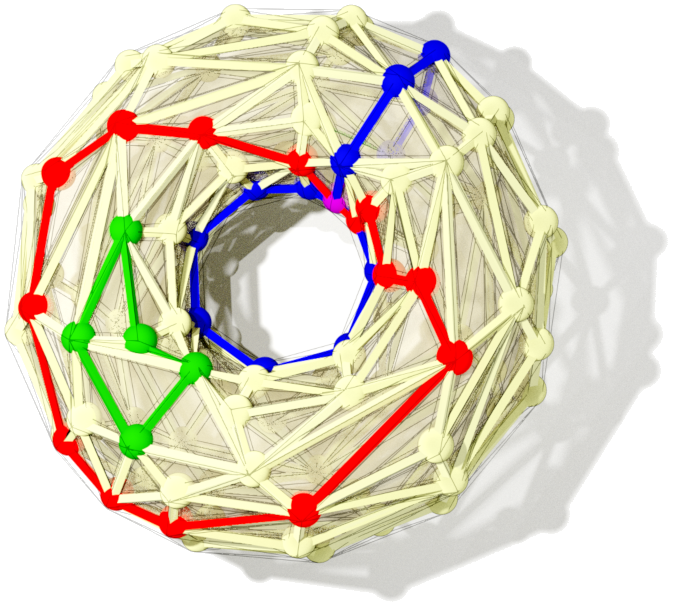}
    \caption{Optimal $\mathcal{H}_1$ homology basis (blue) against an arbitrary cycle homologous to the central loop (red), and a trivial, contractible, boundary cycle (green).}
    \label{fig:homology}
\end{figure}

Each $d$-cycle $g \in \mathcal{Z}_d$ is a formal sum of $d$-simplices satisfying $\partial_d (g)=0$. By assigning weights $w:K_d \rightarrow \mathbb{R}_+$ to these simplices, one can thus define the length of $g$ by adding together weights of its constituent simplices, i.e., $\text{len}(g)=\sum_{\sigma \in g} w(\sigma)$. An {\em optimal} homology basis is the one whose generators have minimum length among all possible bases. Assuming unit, or Euclidean, edge weights, Figure~\ref{fig:homology} depicts the optimal $\mathcal{H}_1$ basis of a torus, in blue, along with a cycle homologous to the generator inscribing the central ``hole", in red, and a trivial, contractible 1-cycle belonging to $\mathcal{B}_1$, in green.

Replacing each matrix $\partial_d$ in \eqref{eq:chcomp} by its transpose $\partial_d^T$, one similarly obtains the $d$-th {\bf cohomology group} of $K$, denoted $\mathcal{H}^d(K;\mathbb{R})$. It is a straightforward consequence of the rank-nullity theorem that there are isomorphisms $\mathcal{H}_d(K;\mathbb{R}) \cong \mathcal{H}^d(K;\mathbb{R})$ between homology and cohomology groups.  


\subsubsection{Hodge Laplacians}

{\em Hodge Laplacians} \cite{horak2013spectra} are to graph Laplacians what simplicial boundary operators are to adjacency matrices.  Given a simplicial complex $K$ and the corresponding sequence \eqref{eq:chcomp}, both composites $\mathcal{L}_d^\text{up} := \partial_{d+1} \partial_{d+1}^T$ and $\mathcal{L}_d^\text{down} := \partial_d^T \partial_d$ furnish linear maps $\mathbb{R}[K_d] \to \mathbb{R}[K_d]$. The $d$-th Hodge Laplacian is their sum:
\begin{align}\label{eq:hodgelapl}
\mathcal{L}_d:=\mathcal{L}_d^\text{up}+\mathcal{L}_d^\text{down}.
\end{align}

An immediate consequence of this definition is that the standard graph Laplacian agrees with the 0-th Hodge Laplacian. The nullity of the graph Laplacian  equals the number of connected components of the underlying graph. Similarly, the kernel of the $d$-th Hodge Laplacian of a simplicial complex $K$ is isomorphic the corresponding $d$-th homology group~\cite{eckmann1944harmonische}:
\begin{align}
    \ker \mathcal{L}_d(K)\cong \mathcal{H}_d(K;\mathbb{R}).
\end{align}

The aforementioned isomorphism still holds for the case of weighted simplicial complexes, where simplices are endowed with non-trivial weights, provided appropriately {\em weighted} Hodge Laplacians~\cite{horak2013spectra} are employed.

\subsection{Graph Neural Networks}

 {\em Graph Neural Networks} (GNNs) provide a general framework for {\em Geometric Deep Learning}~\cite{bronstein2021geometric}, where the input domain, represented by a {\em graph} $G=(V,E)$, is allowed to vary together with the signals that are defined on it. More concretely, the {\em Message Passing Graph Neural Network} (MPGNN) framework generalizes the convolution operation on the edges of a graph $G$ by employing a simple message passing scheme between features of nodes $h_u, u\in V$, and their neighbors $v\in \mathcal{N}_u$. 
 
The output of each layer $\ell$ for each node $u$ can be broadly formulated as:
\begin{align}\label{eq:gnn}
h_u^{\ell+1}=\phi \left( h_u^\ell, \bigoplus_{v\in \mathcal{N}_u} w_{u,v} \cdot \psi \left(h_u^\ell,h_v^\ell\right) \right),
\end{align}
with $\bigoplus$ being a {\em permutation invariant} aggregation, $\phi$ and $\psi$ learnable functions,  and $w_{u,v}$ the weight of edge $(u,v)\in E$. Under this formulation, learnable parameters of $\phi$ and $\psi$ are shared across all nodes in the graph network. 

Each message passing layer with summation aggregation can be described more compactly using matrix notation:
\begin{align}
    H^{\ell+1}=\phi \left( \tilde{L}\psi(H^\ell) \right),
\end{align}
where $\tilde{L}=AWA^T$ is the weighted graph Laplacian matrix, and $H^0$ the $|V| \times F$ matrix of initial node features. This formulation highlights the similarities of GNNs with Laplacian diffusion operations, a fact that we will largely exploit.

The output of a number of message passing iterations results to latent {\em node embeddings}, largely based on the local graph topology at each node. Such embeddings can be subsequently used for node regression, node classification, or, via feature aggregation of all nodes, for graph classification and aggregation tasks.


\section{Related Work}\label{sec:Related}

\subsubsection{Homology Localization} 


The {\em minimum basis problem} in computational topology involves extracting optimal homology generators, with optimality usually expressed in terms of norm or length minimization of cycles. In dimensions exceeding one, this is an NP-hard problem~\cite{chambers2009minimum,chen2011hardness}, whereas the 1-dimensional case succumbs to a polynomial time algorithm~\cite{dey2010approximating, dey2018efficient}. This latter fact spawned a significant body of work examining special cases and computational improvements~\cite{borradaile2016minimum, chen2010measuring, dey2011optimal,erickson2005greedy, Busaryev2011,chen2021decomposition}. 

While the aforementioned methods generally output sets of simplices that form optimal homology generators in their respective class, the rest of the simplices in the complex remain largely oblivious to the location of such optimal cycles in relation to themselves. In our work we attempt to characterize each simplex in the complex with respect to its nearest homology generator, while gaining in efficiency (once the model is sufficiently trained). More similar to our line of work,~\cite{Ebli2019harmonicclustering} implements a homology-aware clustering method for point data.

\subsubsection{Topological Methods in ML}


With the marriage of homology and ML~\cite{Hensel2021TopMLsurvey, love2021topDL, montufar2020can, Hofer2019LearningBarcodes}, it did not take long for GNNs to meet their higher dimensional counterparts in the form of simplicial~\cite{bodnar2021weisfeilerMPNNS, ebli2020simplicial,bunch2020simplicial}, cell~\cite{hajij2021cell, bodnar2021weisfeilerCWNs}, hypergraph~\cite{feng2019hypergraph}, and sheaf~\cite{hansen2020sheaf} neural networks. Most higher dimensional extensions of GNNs aim to operate on the full complex, and redefine the convolution operation in terms of the corresponding Laplacian operator. Contrary to such generalizations, we still operate on a graph. The key difference is that our graph is derived from adjacency and Hodge Laplacian information of the complex at the dimension of interest.

\subsubsection{Pseudoinverse \& Hodge Laplacians in GNNs}

The pseudoinverse and shifted versions of the Laplacian operator are not new in the context of GNNs~\cite{klicpera2019diffusion,wu2019simplifying,alfke2021pseudoinverse}. Nevertheless, they only consider spectral manipulations of the ``classic" graph Laplacian, whose kernel is usually of no practical interest, as long as the graph is connected.

More closely to our work,~\cite{roddenberry2019hodgenet, schaub2018flow} consider edge-flows for signal denoising, interpolation, and source localization based on the {\em linegraph Laplacian}, and the 1-dimensional down Hodge Laplacian $\hL_1^\text{down}$, based on a proxy graph resulting from interchanging edges and nodes. Nevertheless, their analysis remains restricted on graph structures, disregarding any homological features. 

The baseline works considered in the present paper are counterposed in Table~\ref{tab:qualcomp}. The two methods we experimentally compare against, shortloop~\cite{dey2010approximating} and hom\_emb~\cite{chen2021decomposition}, expect complexes embedded in a metric space (first column). Ours can operate purely on the combinatorial structure, and additional structure, such as simplex weights, can be encoded via a weighted Hodge Laplacian, if desired. Although our method depends on training (second column), it is virtually independent of the number of simplices $N$, as long as the complex, or local neighborhoods of simplices, can fit in GPU memory. The reference baseline, shortloop, requires $O(N^4)$ time, whereas hom\_emb requires $O(n_1^{2.37\dots})$ (third column). The alternative baseline considered, distr\_cover\_loc~\cite{Salehi2010Distributed}, does not provide runtime complexity or computation times, but their method relies on $L_1$-relaxation minimization. Finally, post-processing is required to calculate distances to optimal homology generators using the baselines (columns 4-5). In our case, post-processing will be required to identify the generators in terms of the simplices comprising them.

\begin{table}[t]
\setlength{\tabcolsep}{2pt}
\centering
\begin{tabular}{c c c c c c}
    \hline
        & embed & training & time & cycles & dist.     \\
    \hline
    shortloop & yes & no & $O(N^4)$ & yes  & no \\
    distr\_cover\_loc & no & no & N/A &  yes & no  \\
    hom\_emb & yes & no & $O(n_1^{\omega})$ &  yes & no  \\
    ours & no & yes & $O(1)$ & no & yes 
\end{tabular}
\caption{Qualitative comparison of our proposed method against three baselines, shortloop~\cite{dey2010approximating},distr\_cover\_loc~\cite{Salehi2010Distributed}, and hom\_emb~\cite{chen2021decomposition}. $N$ and $n_1$ are the total number of simplices, and edges, respectively, $\omega$ is the matrix multiplication time exponent.}
\label{tab:qualcomp}
\end{table}



\section{Dist2Cycle}\label{sec:Model}

Here we present a model for learning homology-aware distance functions on simplicial complexes.

\subsection{Shifted Inverted Hodge Laplacians} \label{subsec:Ltp}

The spectral properties of the Hodge Laplacian matrices provide salient information regarding the geometry and topology of a simplicial complex, as hinted in Section~\ref{sec:Prelims}. Furthermore, Laplacian flow dynamical systems on simplicial complexes tend to stabilize towards specific spectral regions of the Laplacian~\cite{muhammad2006control}. Nevertheless, the choice of the Laplacian operator with which diffusion is performed impacts greatly the energy distribution on the simplices of interest. 

In Figures~\ref{fig:laplacian_diffusion}\subref{fig:hL1},\subref{fig:hL1_k3},\subref{fig:hL1_3D},\subref{fig:hL1_3D_k5} we perform 12 diffusion steps according to the Hodge Laplacian $\hL_1$ (and its low-rank approximation using the top 3 and 5 eigenpairs, respectively), namely,
\begin{align} \label{eq:diffusion}
\bm{x}_{i+1} =\bm{x}_i + \hL_1 \bm{x}_i,
\end{align} 
with $\bm{x}_0=[1\ 1\ \dots\ 1]^T$ as the initial signal on the 1-simplices of the complex. We show the absolute value of the resulting flow vector at each simplex, on two basic examples. Energy tends to concentrate at well connected simplices, while ignoring the homological features that we are interested in.

The Laplacian diffusion of~\eqref{eq:diffusion} can be seen as a simplified version of a graph convolution that takes place in GNNs~\eqref{eq:gnn}, with all nonlinearities and learnable parameters pruned, and trivial initial features. Thus, in order to focus our attention on optimal homology generators, we must invert the spectrum of the Hodge Laplacian $\hL_1$, while making sure that its kernel will replace the part of the spectrum corresponding to its top eigenvalues. For this purpose we employ a {\em shifted inverted} version of the Hodge Laplacian
\begin{align*}
\Ltpd{d}=(\mathbb{1}+\hL_{d})^{+},
\end{align*}
which makes the $\ker \hL_d$ the prominent part of the spectrum of $\Ltpd{d}$ with eigenvalue 1, onto which the diffusion asymptotically converges. The effect this modified Laplacian matrix has on diffusion is shown in \ref{fig:laplacian_diffusion}\subref{fig:Ltpd1},\subref{fig:Ltpd1_k3},\subref{fig:Ltpd1_3D},\subref{fig:Ltpd1_3D_k5}.

\newcommand{\lgsize}{0.11}
\begin{figure}[t]
     \centering
     \begin{subfigure}[b]{\lgsize\textwidth}
         \centering
         \includegraphics[width=\textwidth]{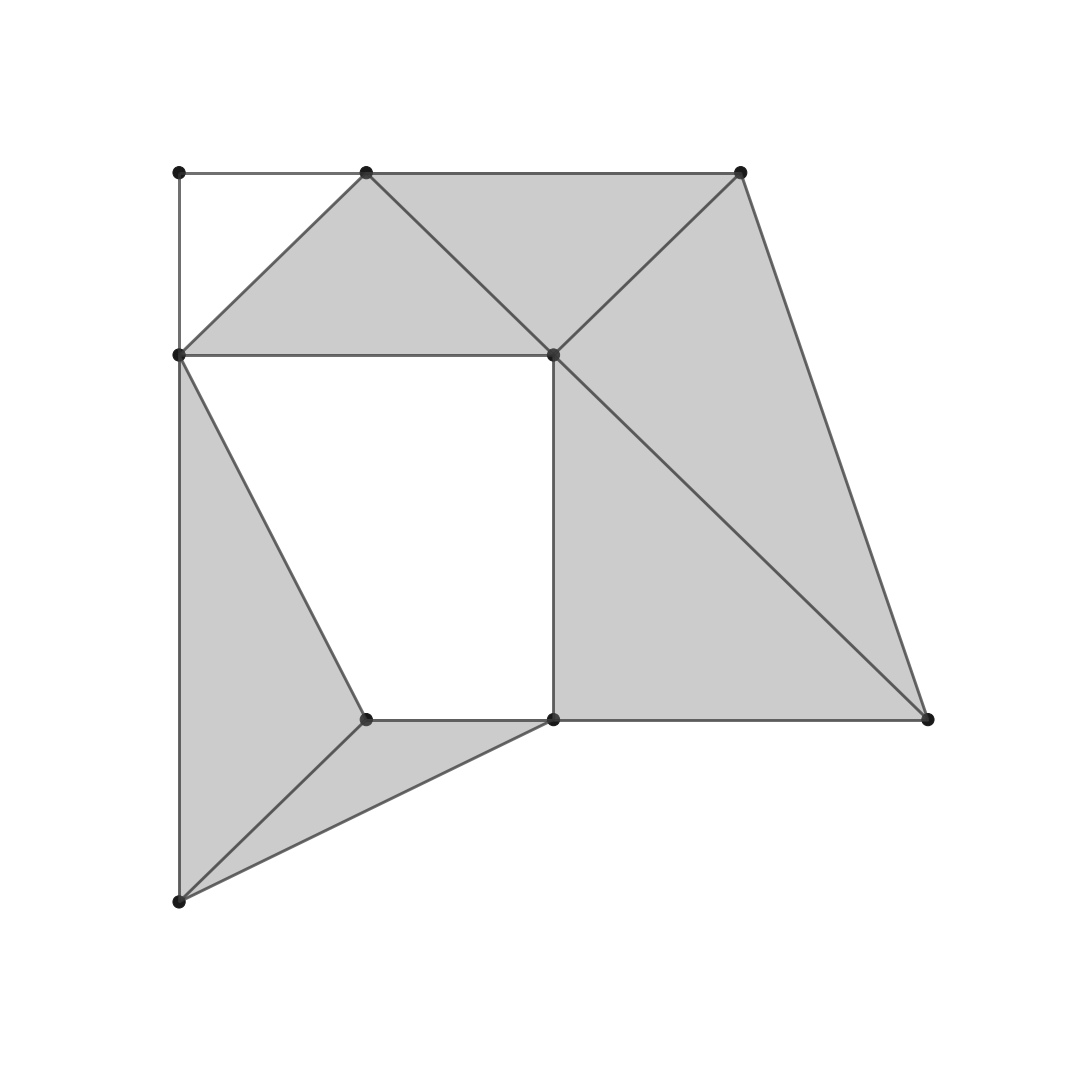}
         \caption{$K$.}
         \label{fig:complex}
     \end{subfigure}
     \hfill
     \begin{subfigure}[b]{\lgsize\textwidth}
         \centering
         \includegraphics[width=\textwidth]{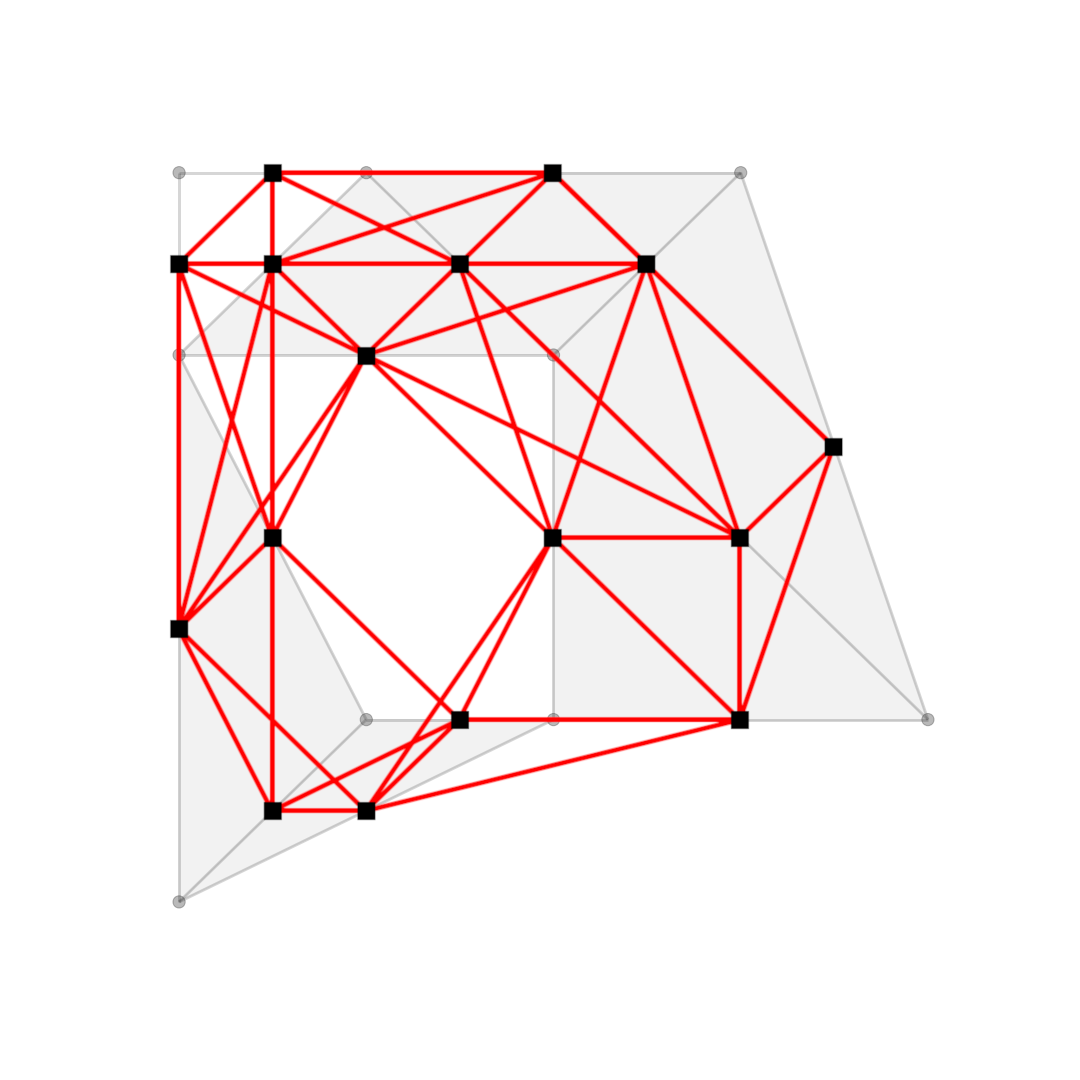}
         \caption{$G_{\hL_1^\text{down}}$.}
         \label{fig:complex_Ldown}
     \end{subfigure}
     \hfill
     \begin{subfigure}[b]{\lgsize\textwidth}
         \centering
         \includegraphics[width=\textwidth]{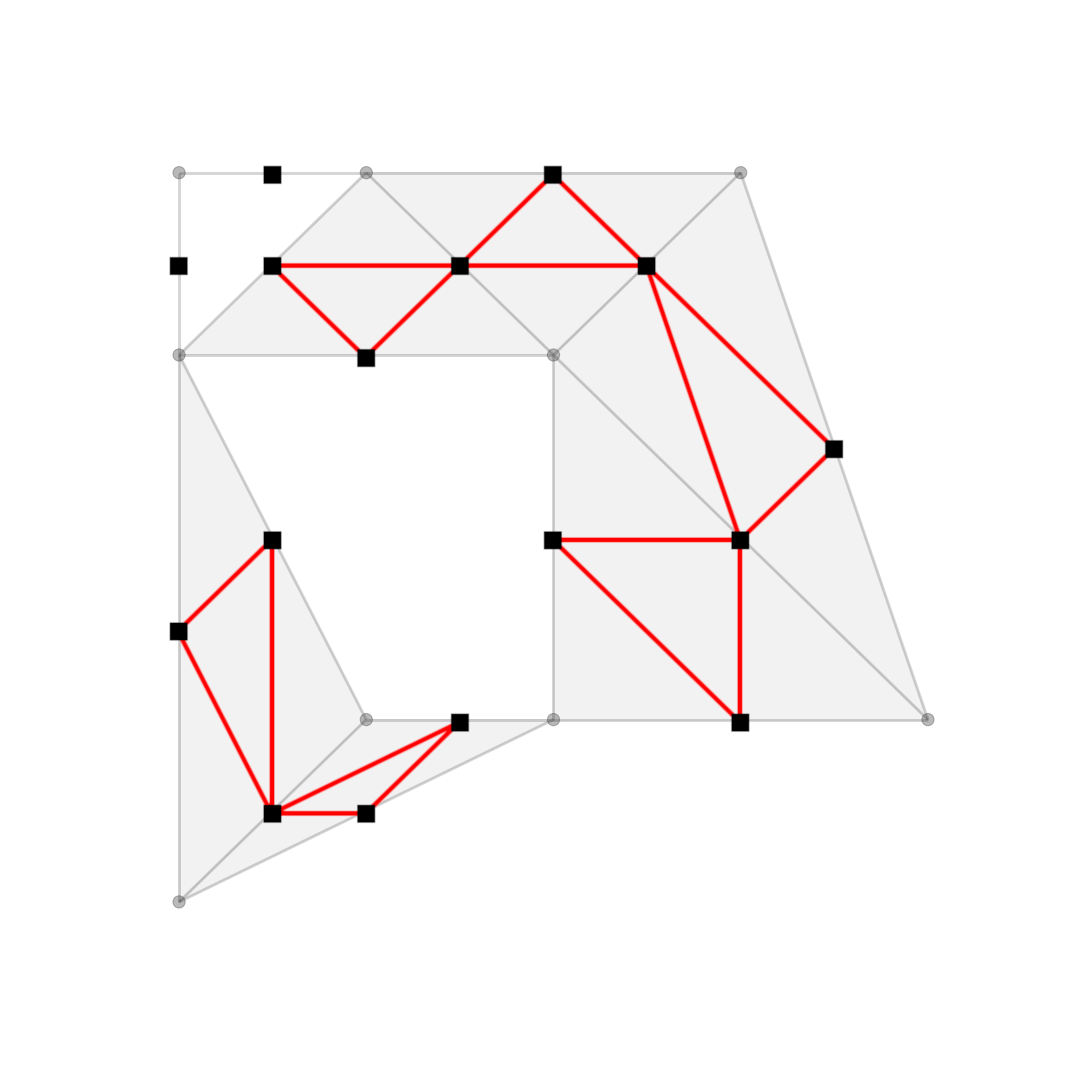}
         \caption{$G_{\hL_1^\text{up}}$.}
         \label{fig:complex_Lup}
     \end{subfigure}
     \hfill
     \begin{subfigure}[b]{\lgsize\textwidth}
         \centering
         \includegraphics[width=\textwidth]{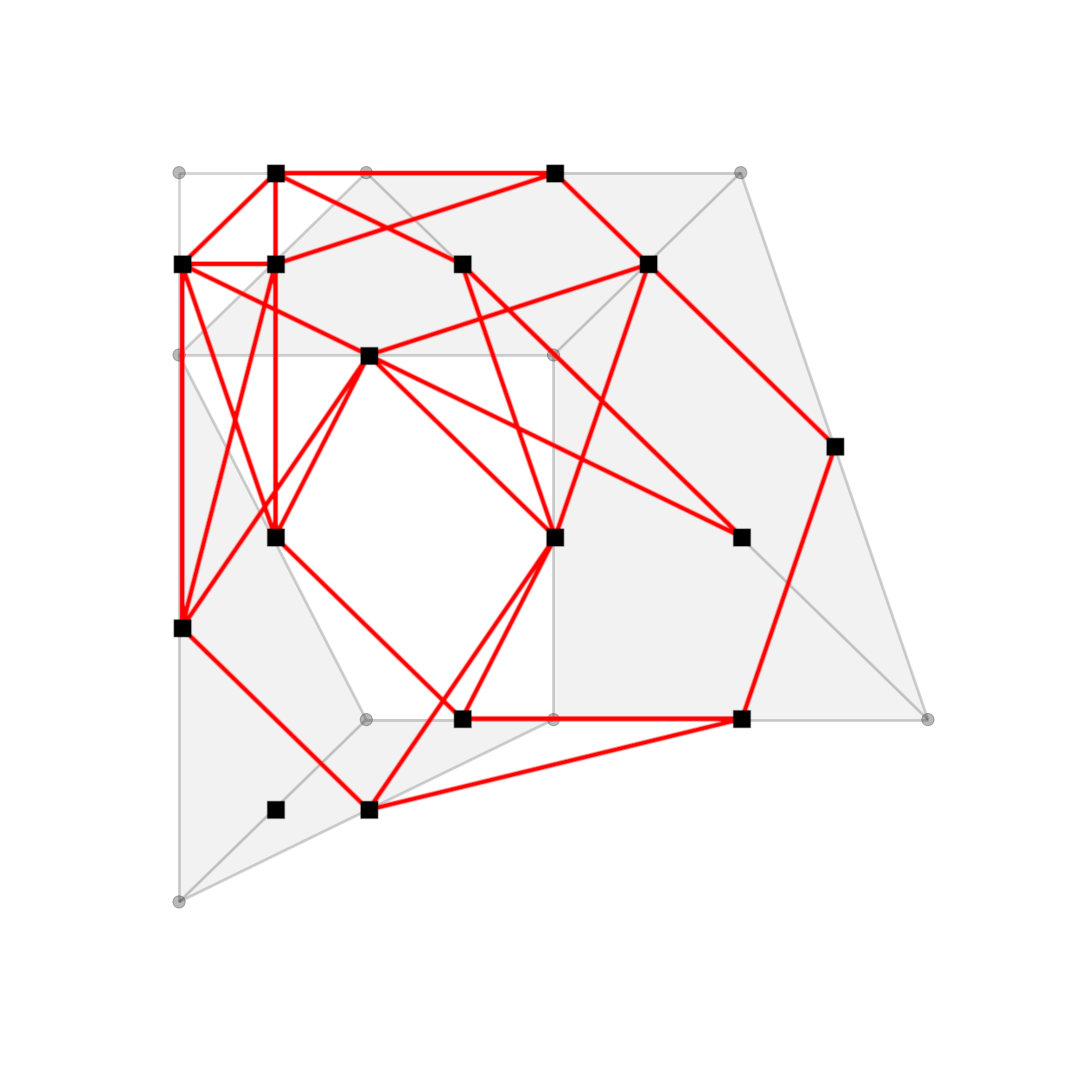}
         \caption{$G_{\hL_1}$.}
         \label{fig:complex_L}
     \end{subfigure}
     \caption{Laplacian graph constructions on example complex $K$ for 1-simplices (square nodes with red edges overlaid on top of the original complex).}
     \label{fig:laplacian_graphs}
\end{figure}

\subsection{From Simplicial Complexes to Graph Neural Networks}\label{subsec:SCGNN}

In order to employ the GNN framework for inference on the complex, we need to express the space of $k$-simplices accordingly. Furthermore, we desire the resulting graph structure to retain the spectral properties of the Laplacian operator of interest. 

We interpret the $|K_d|\times |K_d|$ Hodge Laplacian operator $\hL_d$ (or $\Ltpd{d}$) as the weighted graph Laplacian $\tilde{L}_\text{GNN}=AWA^T$ of a computational graph $G_\text{GNN}=(V_\text{GNN}, E_\text{GNN})$. Under this lens, each $d$-simplex $\sigma$ of the original complex $K$ becomes a node of $G_\text{GNN}$, i.e. $\sigma \in V_\text{GNN}$, and weighted edges are drawn according to the adjacency information encoded in $\hL_d$ ($\Ltpd{d})$. Namely, a weighted edge $(\sigma, \tau)\in E_\text{GNN}$ is drawn between nodes corresponding to $k$-simplices $\sigma$ and $\tau$, whenever the (potentially normalized) Laplacian operator $\hL_d$ ($\Ltpd{d}$) contains a nonzero entry in the respective position. This entry is also used to weight the corresponding edge $w_{\sigma, \tau}=\hL_{\sigma,\tau}$, allowing self-loops. Figure~\ref{fig:laplacian_graphs} provides an example of the resulting graph, what we call the {\em Hodge Laplacian graph}, when using $\hL_1^\text{down}$, $\hL_1^\text{up}$, and $\hL_1$ for extracting adjacency relations on 1-simplices (sans self-loops, for easier vizualization). A somewhat similar approach is followed in~\cite{roddenberry2019hodgenet}, with their mapping akin to the graph in Figure~\ref{fig:complex_Ldown} minus the 2-simplices, as they are only dealing with graphs.

As mentioned in Section~\ref{subsec:Ltp}, we are interested in capturing the spectrum, and thus the connectivity information of $\Ltp$, which is in general a dense matrix and hence computationally prohibitive to work with directly. 
To overcome this issue, we impose the sparsity structure of $\hL$ to $\Ltpd{}$, masking all entries of $\Ltpd{}$ that are zero in the original, sparse, Hodge Laplacian $\hL$. If we denote by $A$ the adjacency matrix encoding the connectivity of $\hL$, with
\[
A_{u,v} = \begin{cases} 1 & \text{ if } \hL_{u,v}\neq 0, \text{ and} \\
                        0 & \text{ otherwise}
        \end{cases},
\]
this can be achieved with the Hadamard product $A \odot \Ltp$. The resulting graph is called the {\bf Hodge Laplacian graph} throughout this paper.

While more sophisticated methods for spectral sparsification exist~\cite{spielman2011graph}, imposing the connectivity dictated by $\hL$ or $\hL^\text{down}$ seems to preserve all important adjacency information required for the task at hand, while not annihilating important spectral information. Furthermore, in the context of learning, this approach is reminiscent to inference with missing values, which GNNs are known to handle well~\cite{Jiaxuan2020missing}.

\subsection{Shifted Inverted Laplacian GNNs for Homology Localization} \label{subsec:SFGNN}

We are now ready to propose a {\em Simplicial Neural Network} model for homology localization. By following the construction described in Section~\ref{subsec:SCGNN} we obtain a weighted computational graph $G_\text{GNN}=(V_\text{GNN}, E_\text{GNN})$, with weights according to $\Ltp$ and adjacency dictated by $\hL$ (or $\hL^\text{down}$). Thus, graph convolution (message passing) on the $G_\text{GNN}$ can be summarized as:
\begin{align} \label{eq:SFGC}
    H^{\ell+1}=\phi \left( A \odot \Ltpd{d} H^\ell W^\ell \right),
\end{align}
where $A$ is the adjacency matrix describing the selected sparsification regime according to $\hL$ (or $\hL^\text{down}$), $\Ltpd{d}$ the {\em shifted inverted Hodge Laplacian} in dimension $d$ (Section~\ref{subsec:Ltp}), and $\odot$ denoting the Hadamard product. The learnable weights of the model at layer $\ell$ are denoted as $W^\ell$, and $\phi$ can be any activation function, such as ReLU, Sigmoid, etc. Finally, $H^\ell$ is the $|V_\text{GNN}|\times F$ feature matrix having the $F$-dimensional features of each node (i.e. $d$-simplex) as rows.  

To aid the task of homology localization, we encode both local and global information at each node. Locality is incorporated by computing betti numbers $[\beta_0, \dots, \beta_{d+1}]$ of the {\em link} at each $d$-simplex $\sigma$  --- this is the subcomplex consisting of all simplices $\tau$ for which $\sigma \cap \tau$ is empty and $\sigma \cup \tau$ is a simplex in $K$. Global features manifest in the form of spectral embeddings of the $d$-simplices in the space spanned by singular vectors corresponding to the largest $k$ singular values of $\Ltpd{d}$. Denoting the appropriately permuted singular value decomposition (SVD) of $\Ltpd{d}$ as $\Ltpd{d}=U\Sigma V^T$ with $\Sigma$ containing in its diagonal the singular values of $\Ltp$ in descending order, the rows of the matrix $U_{1:k}$ formed by the first $k$ singular vectors constitute coordinates of the simplices in the eigenspace of $\Ltpd{d}$. Due to the shift-invert operation of $\Ltp$, this scheme effectively embeds the $d$-simplices in the spectral subspace corresponding to the kernel of $\hL_{d}$, i.e. the space encoding homological information. 

\begin{figure*}[ht]
     \centering    
     \def\arraystretch{0.0}
     \begin{tabular}{@{}c@{}c@{}c@{}c@{}}
     \begin{subfigure}[b]{\errorplt\textwidth}
         \centering
         \includegraphics[width=\textwidth]{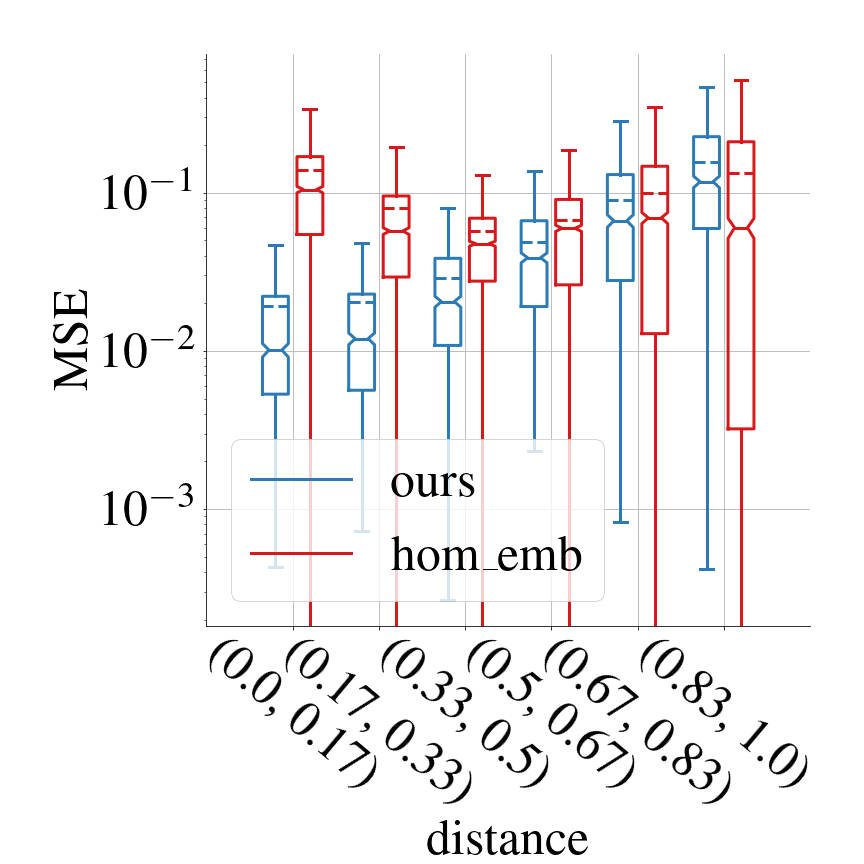}
         \label{fig:2DMSE}
     \end{subfigure} &
     \hfill
     \begin{subfigure}[b]{\scaleplt\textwidth}
         \centering
         \includegraphics[width=\textwidth]{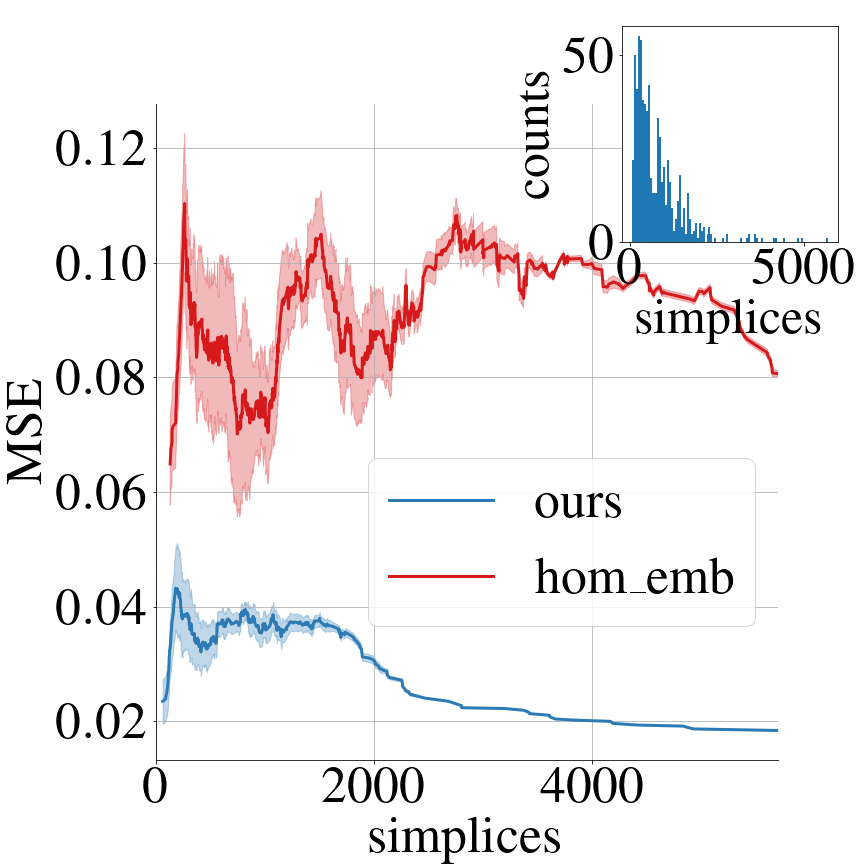}
         \label{fig:2Dsimplices}
     \end{subfigure}&
     \hfill
     \begin{subfigure}[b]{\scaleplt\textwidth}
         \centering
         \includegraphics[width=\textwidth]{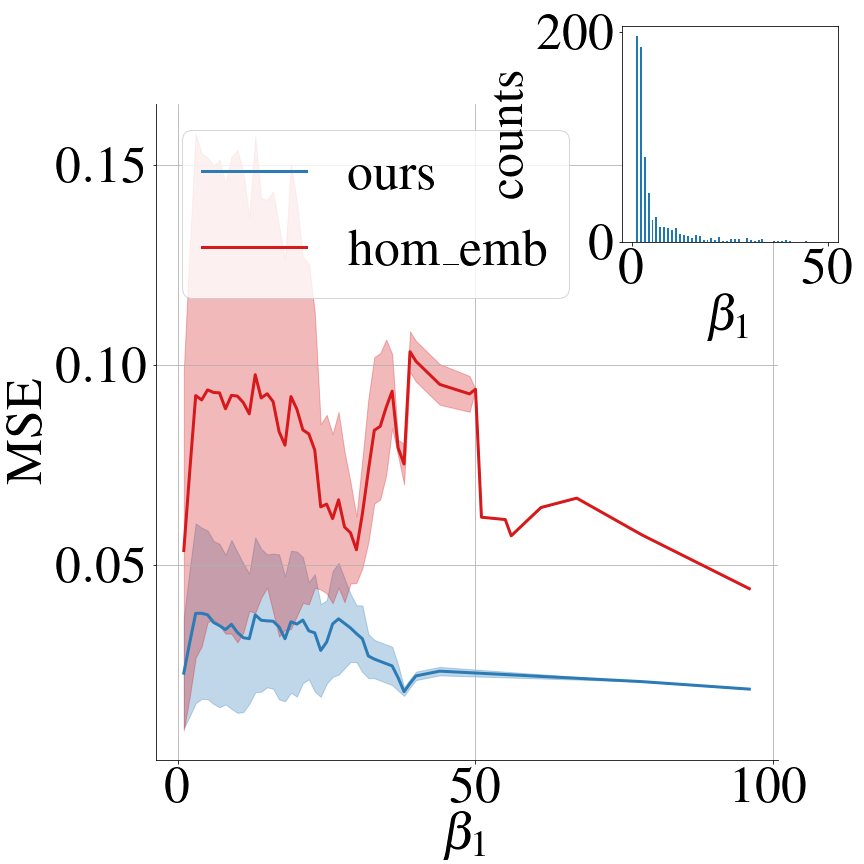}
         \label{fig:2Dbetti}
     \end{subfigure} &
     \hfill
     \begin{subfigure}[b]{\scaleplt\textwidth}
         \centering
         \includegraphics[width=\textwidth]{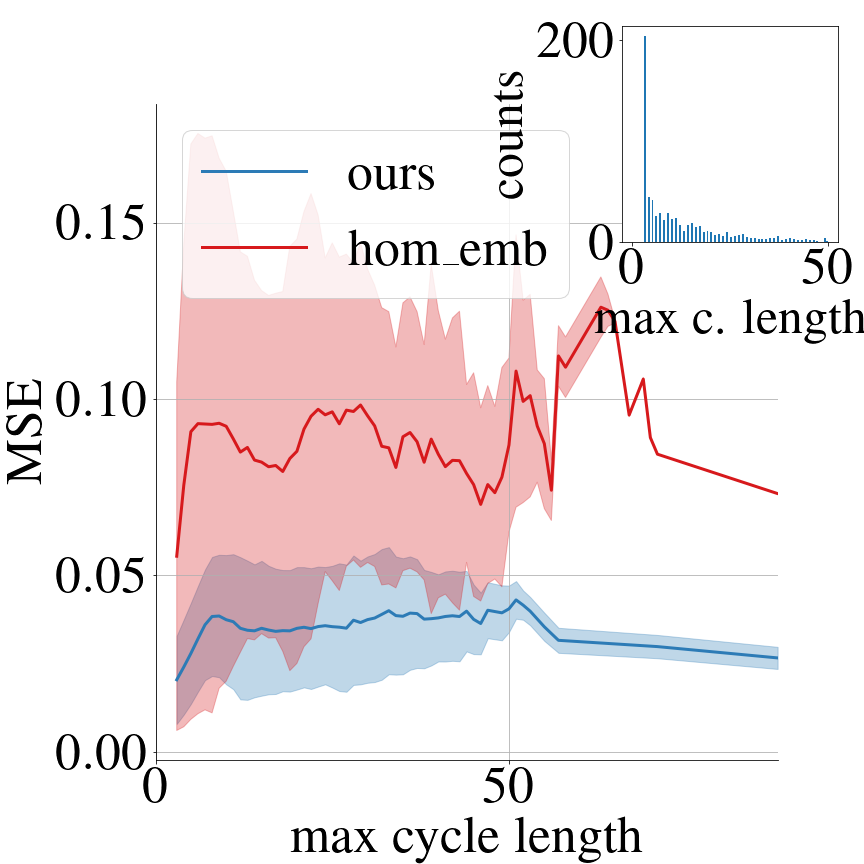}
         \label{fig:2DmaxCycle}
     \end{subfigure} \\
     \begin{subfigure}[b]{\errorplt\textwidth}
         \centering
         \includegraphics[width=\textwidth]{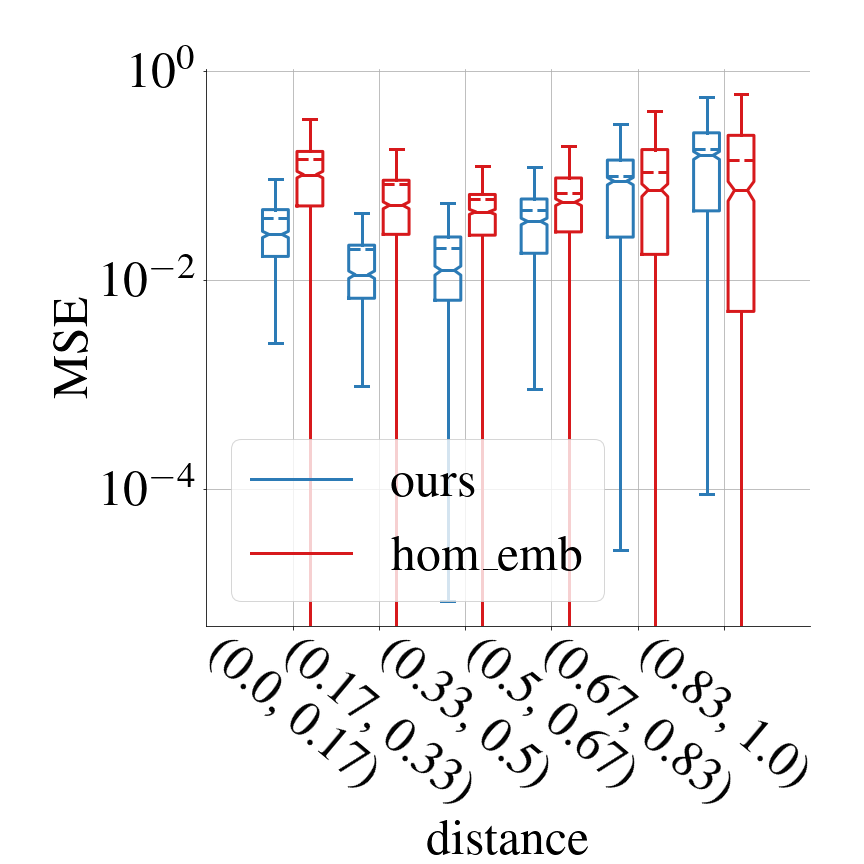}
         \label{fig:3DMSE}
     \end{subfigure} &
     \hfill
     \begin{subfigure}[b]{\scaleplt\textwidth}
         \centering
         \includegraphics[width=\textwidth]{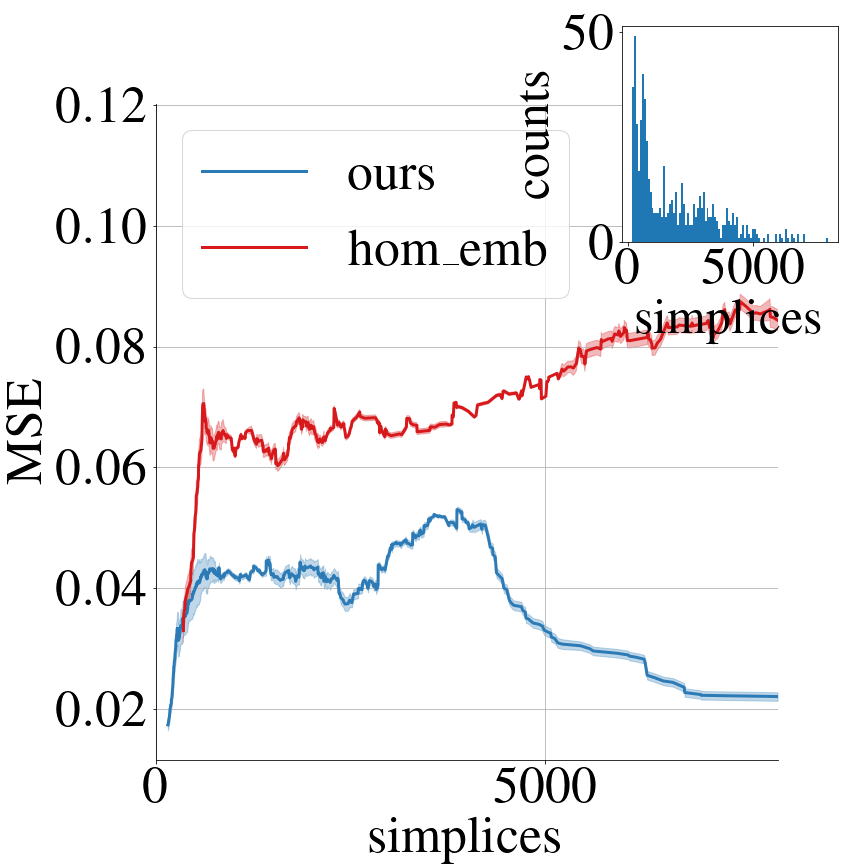}
         \label{fig:3Dsimplices}
     \end{subfigure} &
     \hfill
     \begin{subfigure}[b]{\scaleplt\textwidth}
         \centering
         \includegraphics[width=\textwidth]{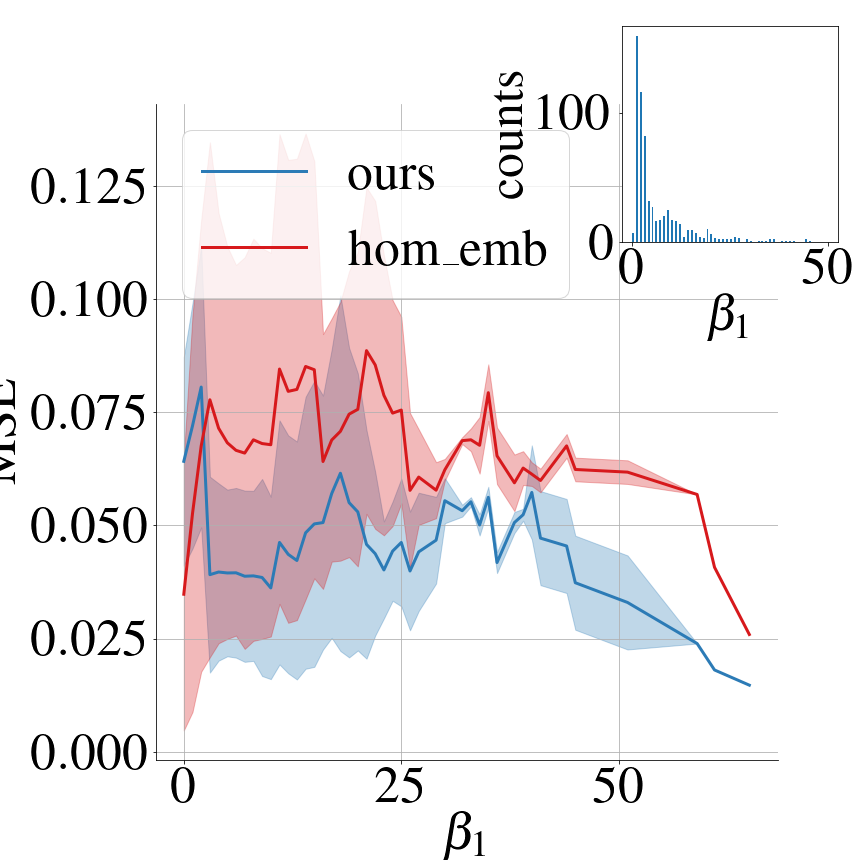}
         \label{fig:3Dbetti}
     \end{subfigure} &
     \hfill
     \begin{subfigure}[b]{\scaleplt\textwidth}
         \centering
         \includegraphics[width=\textwidth]{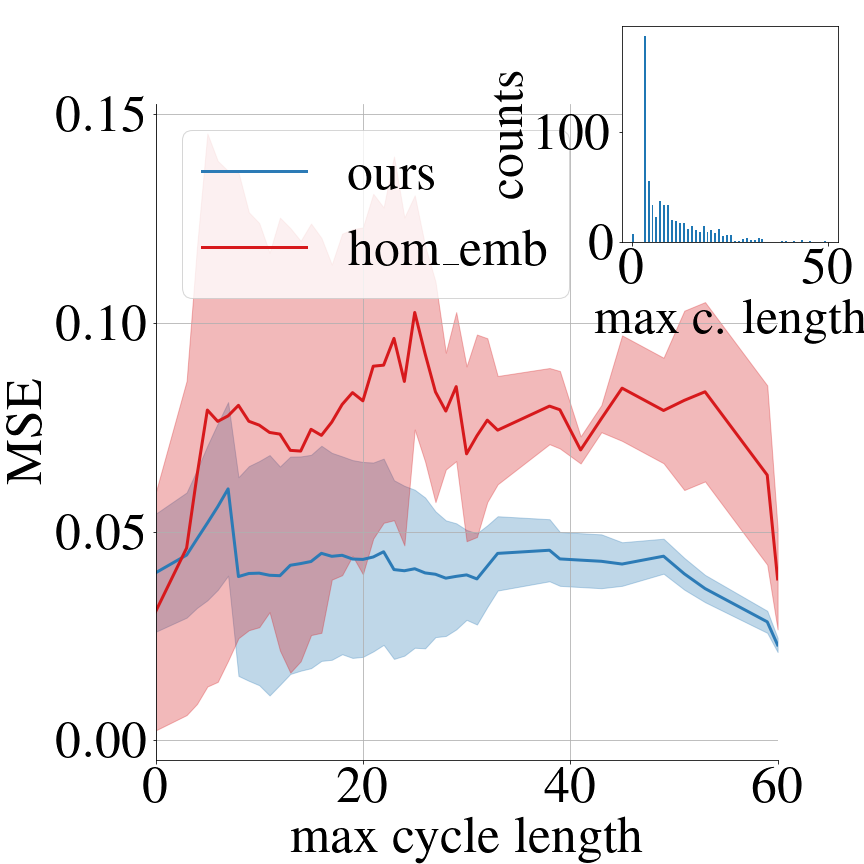}
         \label{fig:3DmaxCycle}
     \end{subfigure}
     \end{tabular}
     \caption{MSE error plots comparing our method against hom\_emb for the \texttt{TORI} dataset in 2D (top) and 3D (bottom). Due to hardness of computing a ``ground truth", the best known baseline appoximation, shortloop, acts as reference. Column 1 shows MSE against stratified distance values (x-axis). Columns 2, 3 and 4 plot MSE against the number of simplices, the homology rank $\beta_1$, and the maximum cycle length, respectively. Dashed line: mean, solid horizontal line: median, box limits: quartiles, whiskers: error range, shaded regions: standard deviation, insets: histograms for the respective parameters in the test set.}
     \label{fig:2D3Derrors}
\end{figure*}
\section{Evaluation}\label{sec:Experiments}
In this section we describe the function learned by our model, the dataset we developed to train and evaluate our model (Section~\ref{subsec:dataset}), the experiments conducted (Section~\ref{subsec:setting}) and an analysis of the results (Section~\ref{subsec:results})\footnote{Code \& models: \url{https://github.com/alexdkeros/Dist2Cycle}}.

\subsection{Nearest Optimal Homology Generator}

Our model approximates a function that depends on the optimal homology generators. To validate our model we only consider  1-simplices, since efficient algorithms acting as proxies to ground truth, and combinatorial, baseline, methods only exist for $d=1$. We  denote by $\mathcal{Q}_1$ the set of optimal generators of $\mathcal{H}_1$. For each simplex $\sigma \in K_1$ we seek to learn its distance from the nearest optimal $g\in \mathcal{Q}_1$,
\begin{align}\label{eq:distTocycle}
    f(\sigma)=\min_{g\in \mathcal{Q}_1} \hat{d}(\sigma,g).
\end{align}
As distance $d(\sigma, g)$ between a $k$-simplex $\sigma$ and a $k$-dimensional homology generator we consider the minimum number of $k$-simplices required to reach any $k$-simplex $\rho \in g$ participating in a $k$-cycle $g$. To keep the function complex-independent, we then normalize the distance in the range $[0,1]$, obtaining $\hat{d}(\cdot)$, with simplices near an optimal homology generator attaining values close to zero.

\begin{figure*}[t]
    \centering
    \begin{tabular}{@{}c c@{}@{}c@{}@{}c@{}@{}c@{}@{}c@{}@{}c@{}@{}c@{}@{}c@{}@{}}
        \rotatebox{90}{Ours} \mbox{\vline height 10ex}&
         \includegraphics[width=\cmplxx\textwidth]{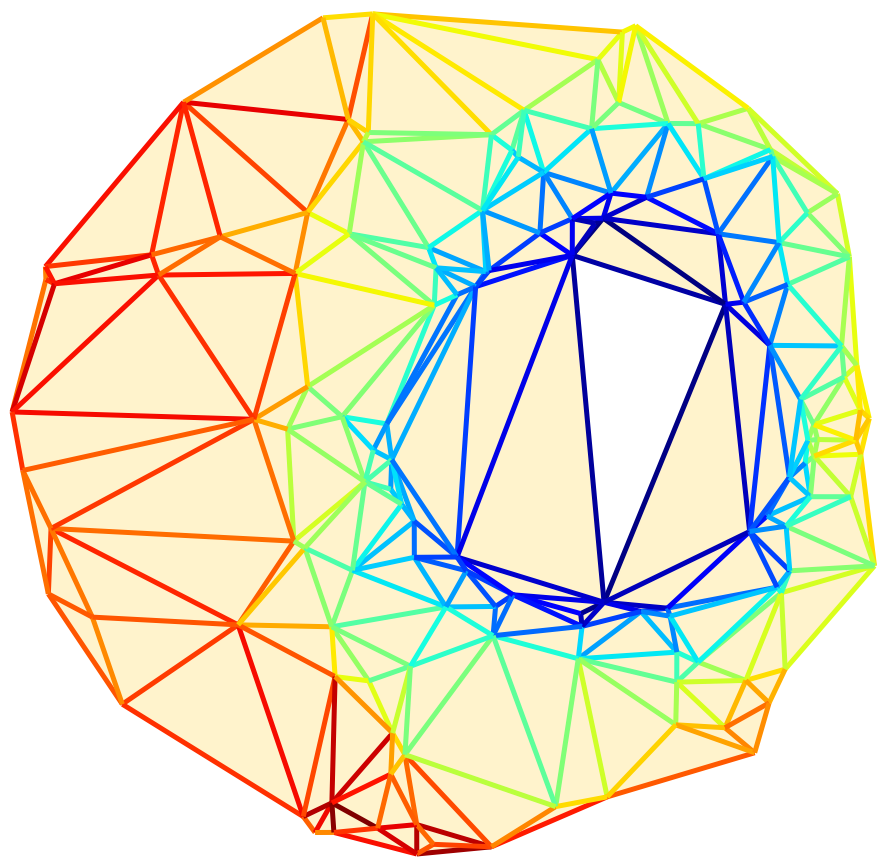} &          \includegraphics[width=\cmplxx\textwidth]{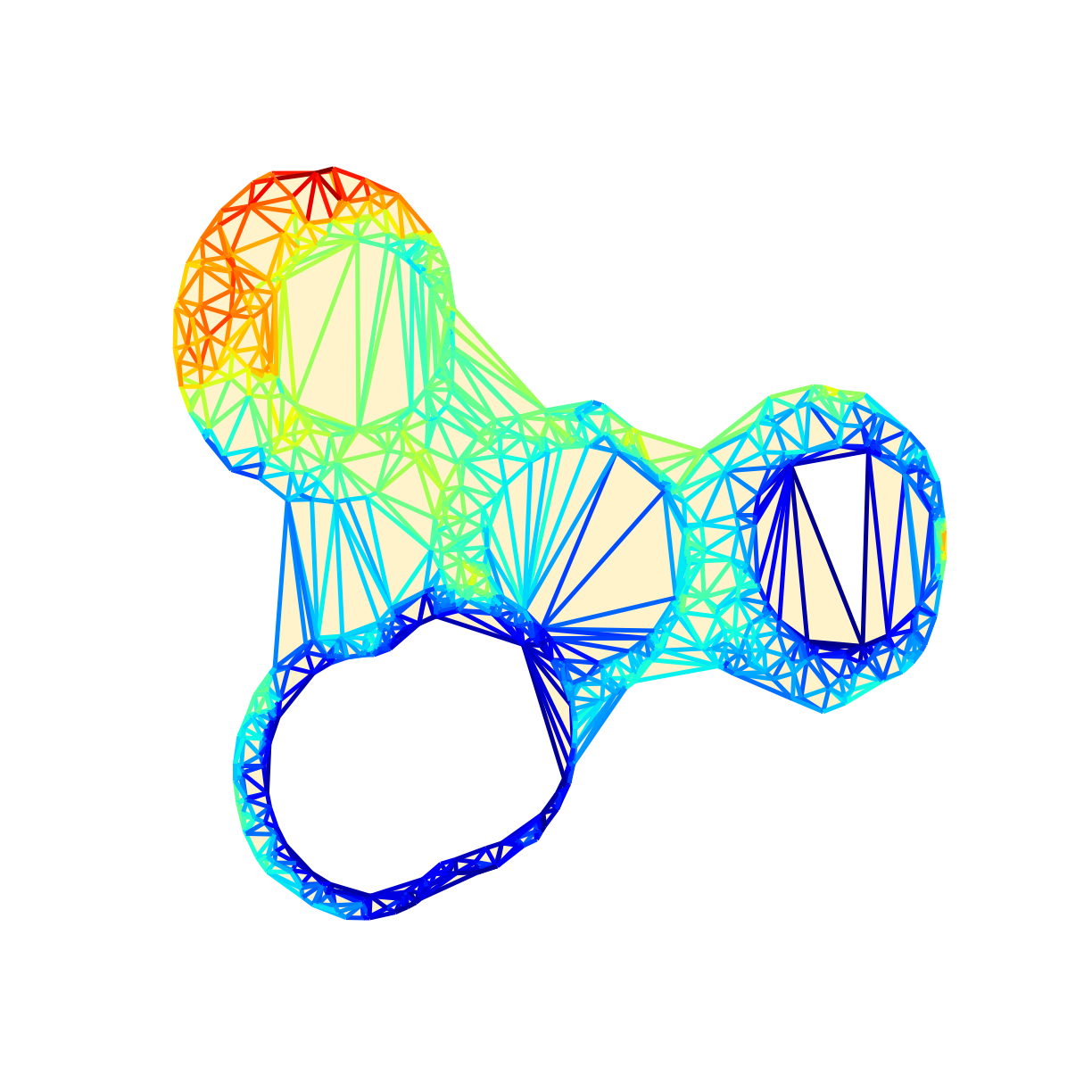} &          \includegraphics[width=\cmplxx\textwidth]{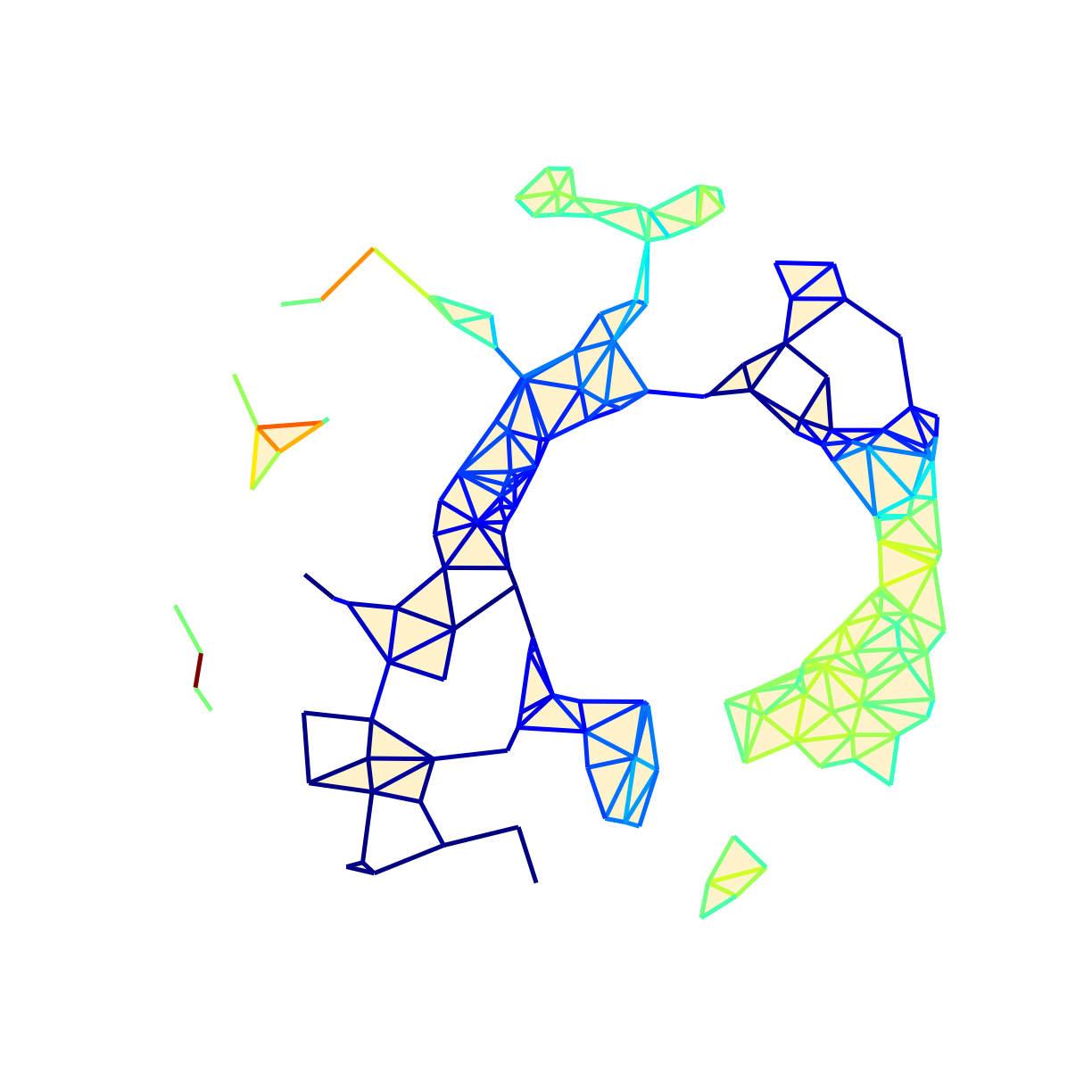} &          
         \includegraphics[width=\cmplxx\textwidth]{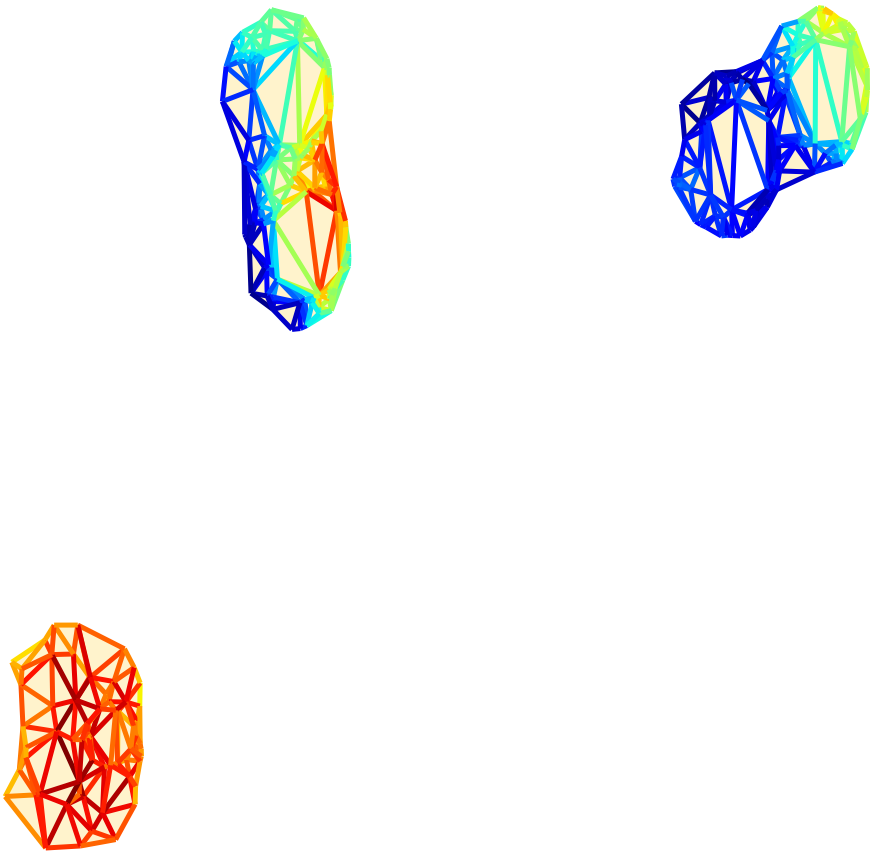} &             \includegraphics[width=\cmplxDD\textwidth]{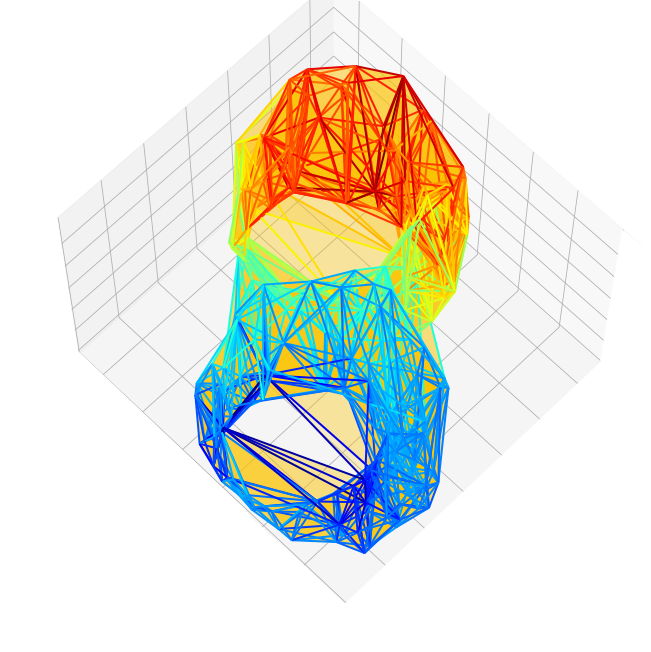} &          \includegraphics[width=\cmplxDD\textwidth]{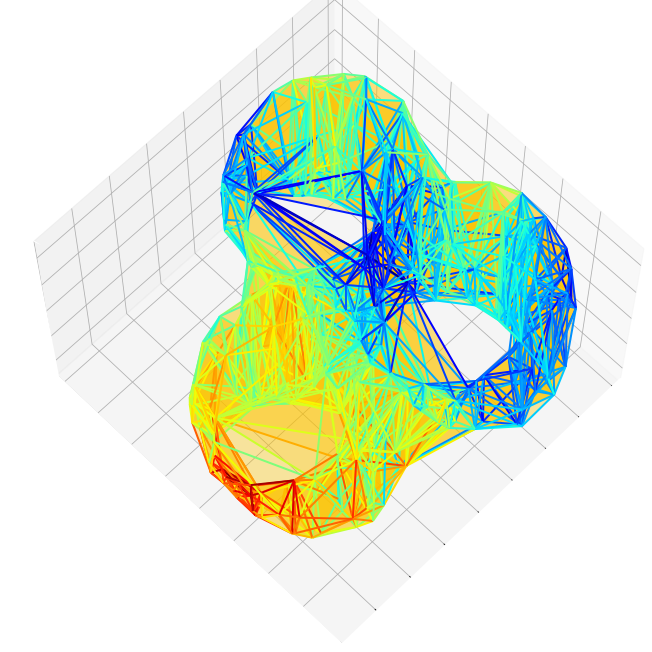} &          \includegraphics[width=\cmplxDD\textwidth]{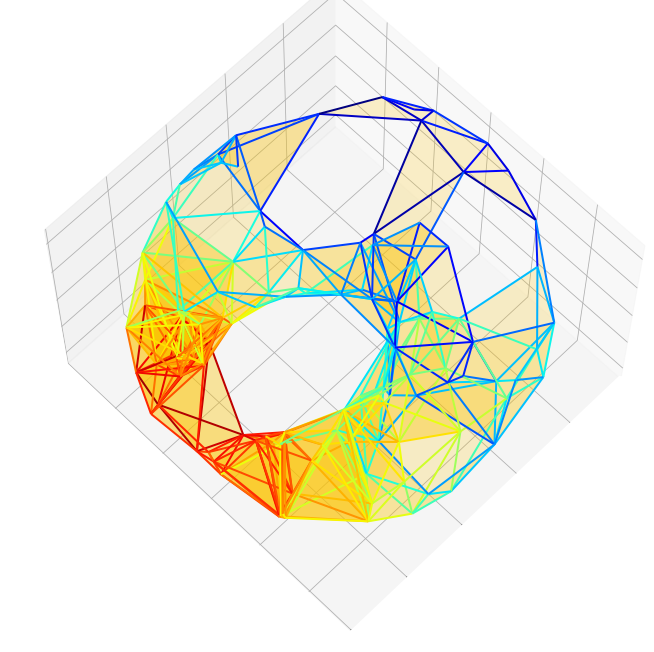} &          \includegraphics[width=\cmplxDD\textwidth]{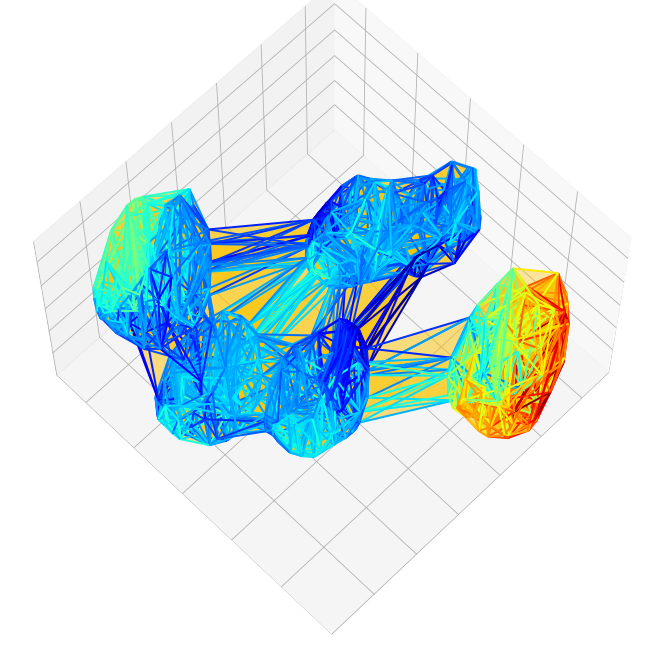}\\
         \rotatebox{90}{Reference} \mbox{\vline height10ex}&
         \includegraphics[width=\cmplxx\textwidth]{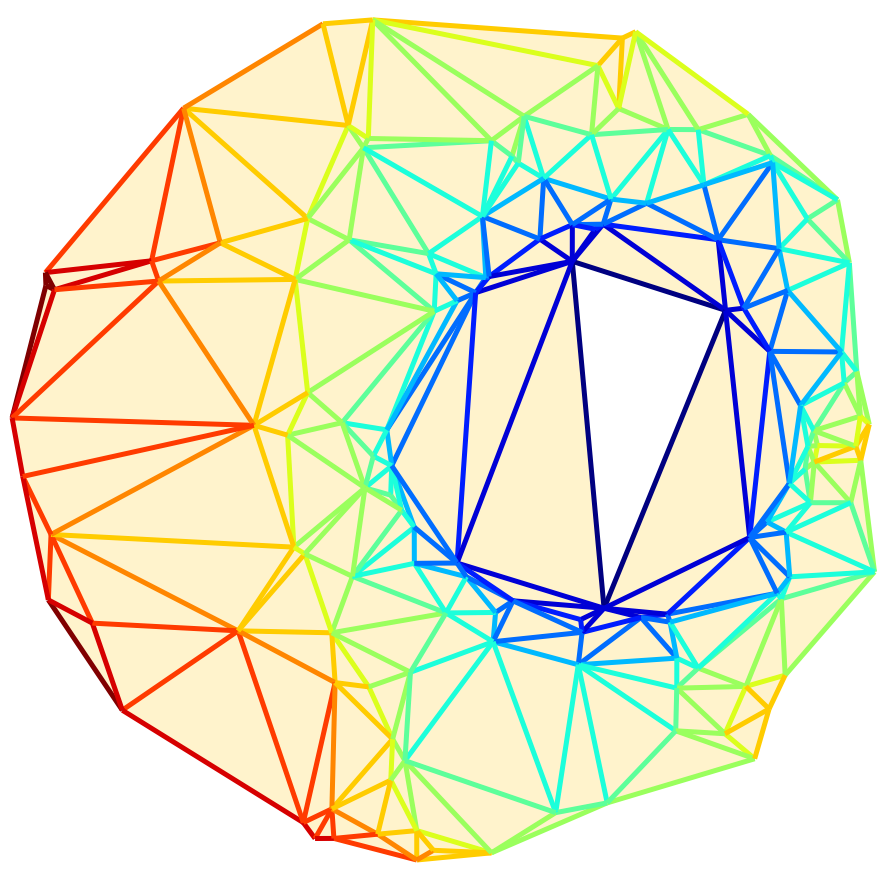} &         \includegraphics[,width=\cmplxx\textwidth]{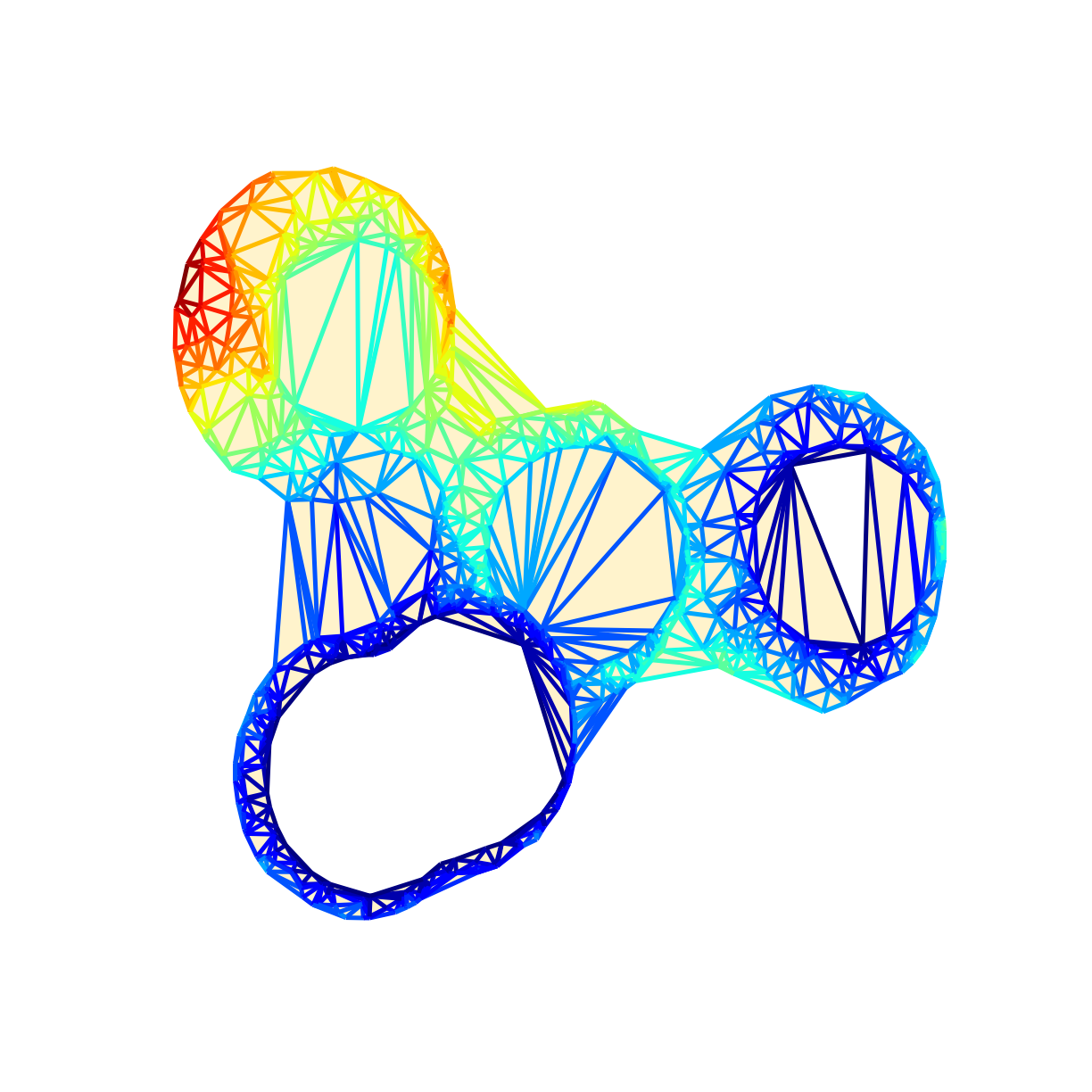} &         \includegraphics[width=\cmplxx\textwidth]{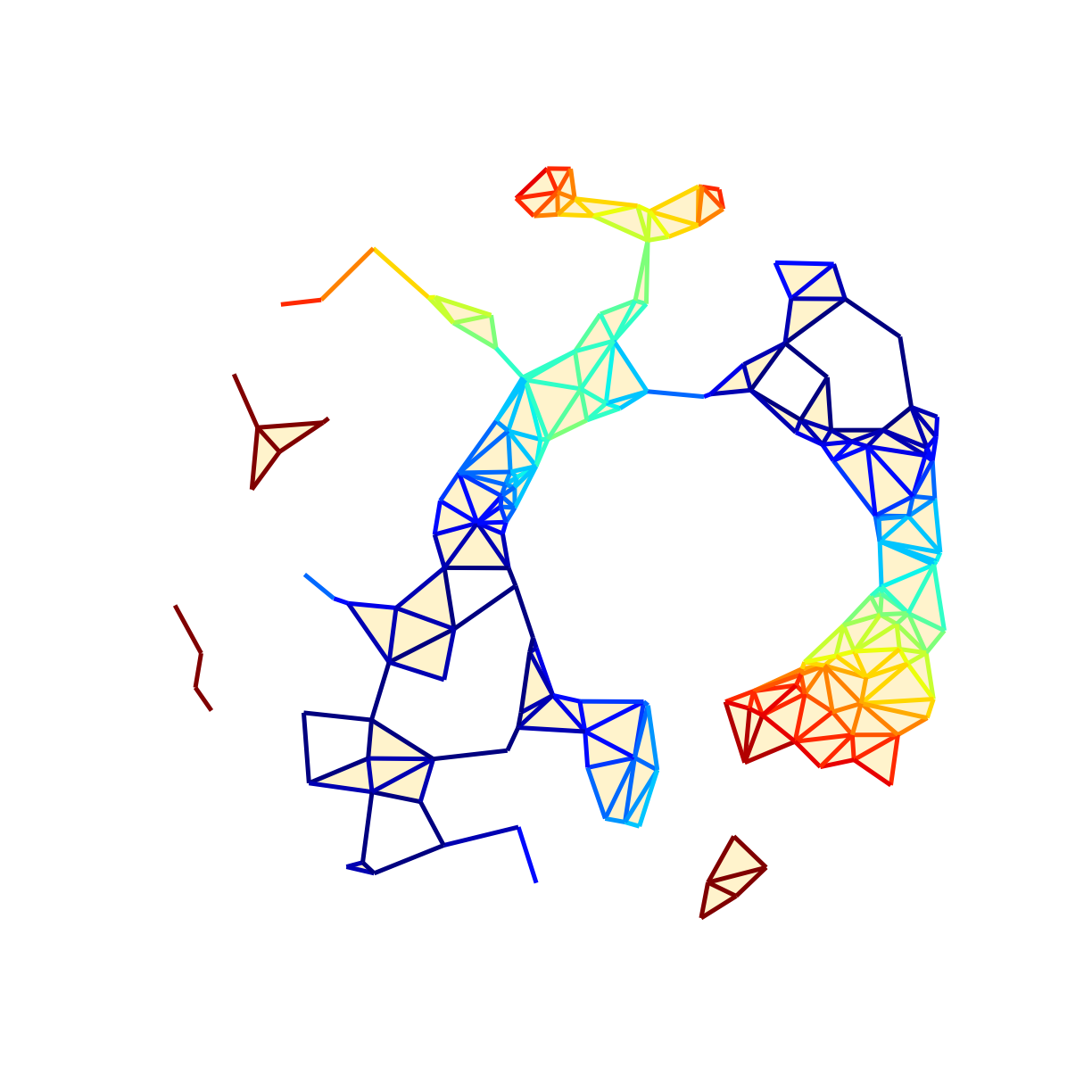}  &         
         \includegraphics[width=\cmplxx\textwidth]{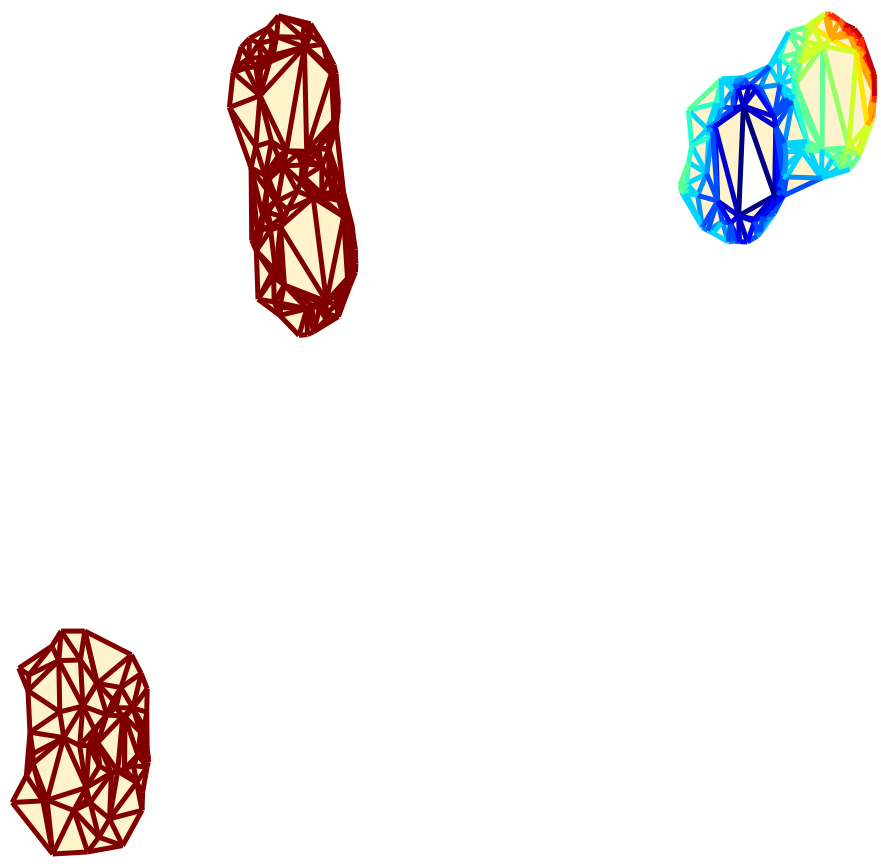} &
         \includegraphics[width=\cmplxDD\textwidth]{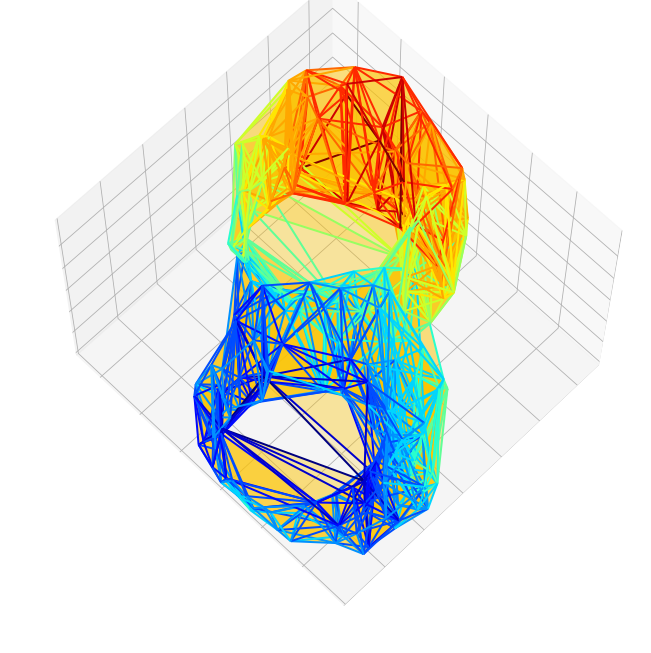} &         \includegraphics[,width=\cmplxDD\textwidth]{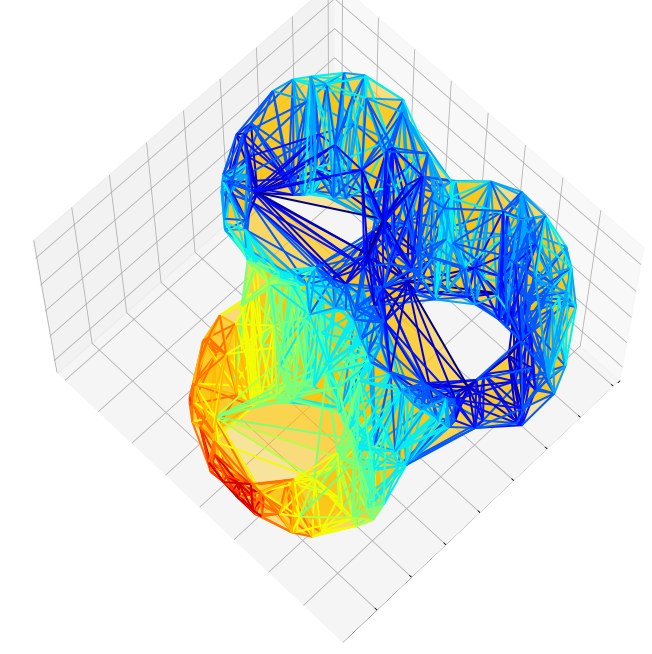} &         \includegraphics[width=\cmplxDD\textwidth]{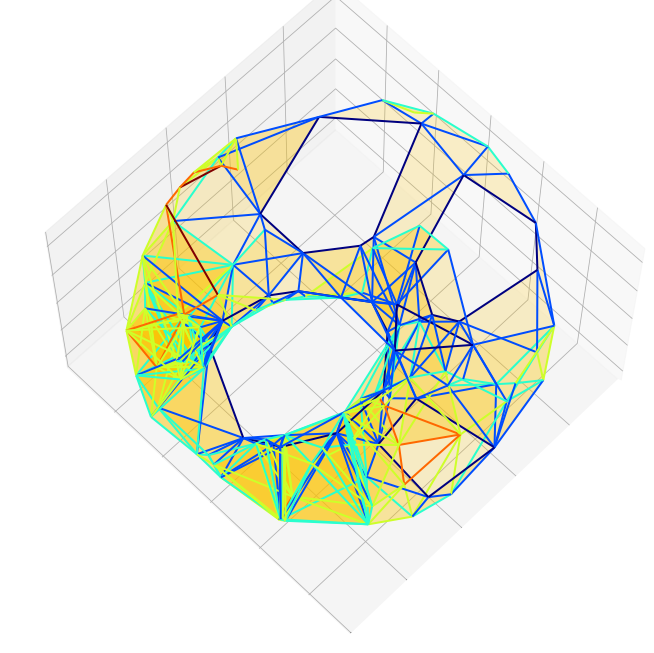}  &         
         \includegraphics[width=\cmplxDD\textwidth]{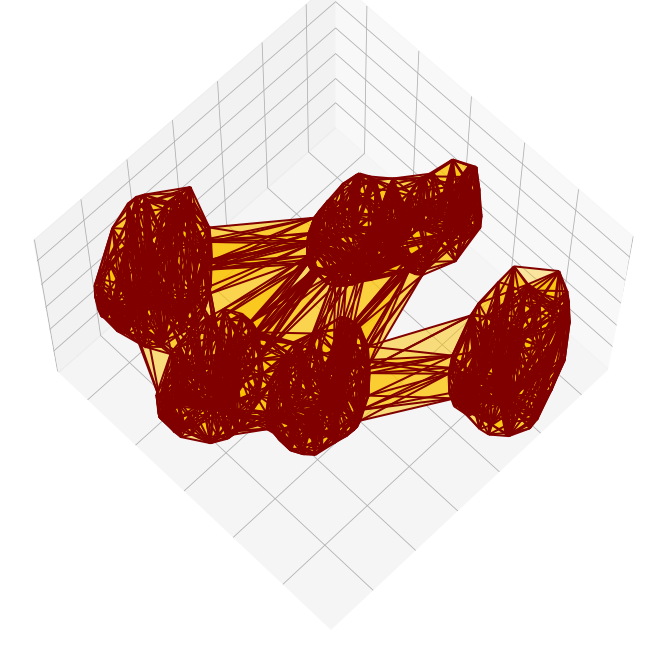}\\
    \end{tabular}
\caption{Qualitative comparisons of selected complexes from the 2D \texttt{TORI} (four left) and 3D (four right) test set. The 1-simplices of the complexes are color-coded according to their distance from the nearest homology cycle, with blue indicating close proximity to an optimal homology cycle, and red indicating large distance from a homology cycle. }    \label{fig:complviz}
\end{figure*}

\begin{figure*}[t]
    \centering
    \begin{tabular}{cccc} 
        \includegraphics[width=\cmplx\textwidth]{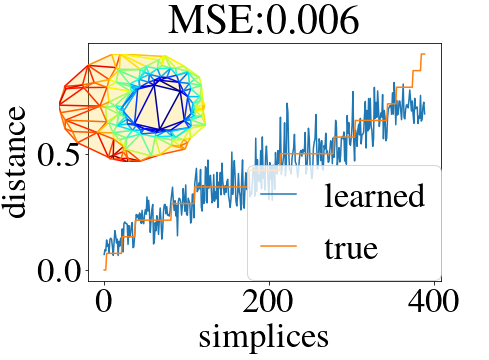} &  
        \includegraphics[width=\cmplx\textwidth]{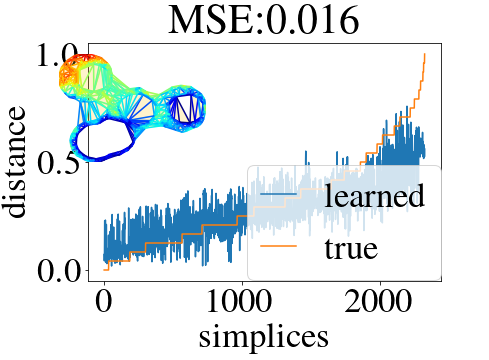}&
        \includegraphics[width=\cmplx\textwidth]{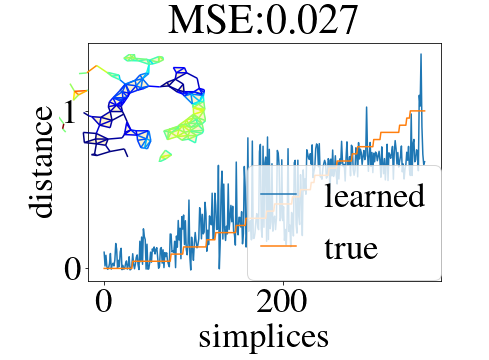}&
        \includegraphics[width=\cmplx\textwidth]{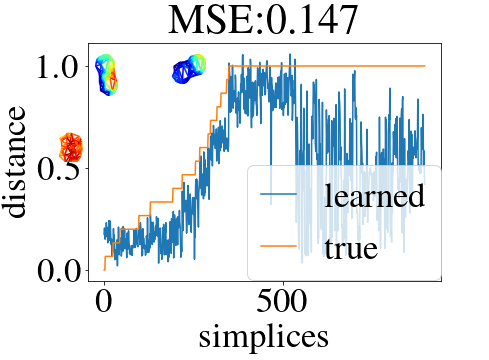}
        \\
             \includegraphics[width=\cmplxD\textwidth]{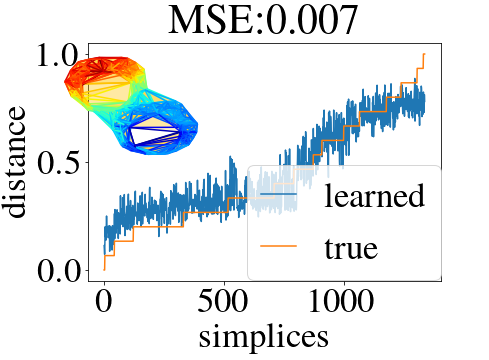} &  
        \includegraphics[width=\cmplxD\textwidth]{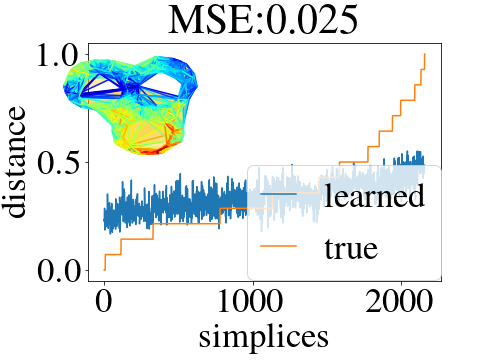}&
        \includegraphics[width=\cmplxD\textwidth]{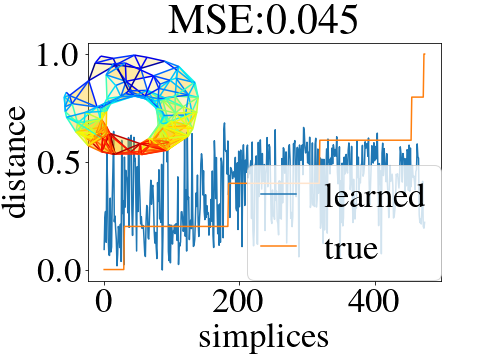}&
        \includegraphics[width=\cmplxD\textwidth]{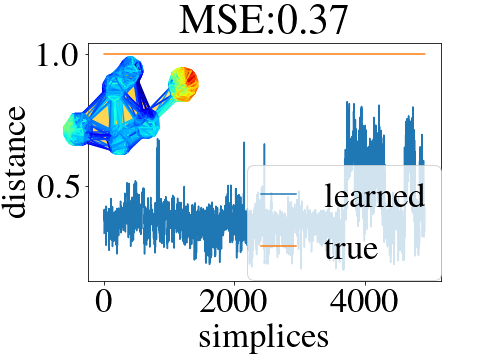}
    \end{tabular}
    \caption{Plots of predicted distances (blue) of the simplices sorted in increasing ground truth distance value (orange) for the Tori dataset in 2D (top) and 3D (bottom). The plots show four cases (columns) ranging from the best (left) to the worst (right). }
    \label{fig:complplots}
\end{figure*}

\subsection{\texttt{TORI} Dataset}\label{subsec:dataset}
Our \texttt{TORI} datasets consist of Alpha complexes~\cite{edelsbrunner2010alpha} that originate from considering ``snapshots" of filtrations~\cite{edelsbrunner2010computational} on points sampled from tori manifolds of diverse topological characteristics, in 2 and 3 dimensions. We seek to capture richness of homological information,  controlability in terms of scalability in the number of simplices and homology cycles, as well as ease of visualization.

We first sampled 400 point clouds from randomly generated configurations of tori and pinched tori, with number of ``holes" ranging from 1 to 5, to which Gaussian noise is added. We then constructed Alpha filtrations on the collection of point clouds, i.e. sequences of simplicial complexes dictated by a monotonically increasing distance parameter $\alpha$. Tracking homological changes in the sequence of complexes results in a {\em barcode} representation, with one bar per homology feature, that spans a range of $\alpha$ values. The longer the bar, the more persistent, and possibly ``significant", a homological feature is. From these barcodes we considered the 5 most persistent features, expressed by the longest bars. The birth and death values of these features were deemed as appropriate points to capture ``snapshots" of the complexes, guaranteed to contain interesting large and small scale homological information. This pipeline yields a collection of 2000 complexes for each dataset. The generality of the datasets stems from the spurious homological features occuring while considering filtrations of noisy point clouds, as confirmed by the histogram insets describing the test sets in Figure~\ref{fig:2D3Derrors}. The number of simplices ranges from tens to thousands, and betti numbers, i.e. number of homology cycles, from 0 up to 66. Furthermore, homology generators present in the dataset can contain from 3 up to 60 1-simplices.

Unfortunately, constructing a general enough dataset that contains rich homological information, whilst explicitly knowing ground-truth optimal homology generators, is not feasible. Thus, we calculate the reference function using optimal homology generators of $\mathcal{H}_1$ discovered via {\em shortloop}~\cite{dey2010approximating}, by calculating the normalized ``hop" distance of each simplex to its nearest optimal generator (see Eq.\eqref{eq:distTocycle}). Thus, shortloop algorithm acts both as a baseline, and a \emph{ground truth proxy}.

\begin{figure}[h]
     \centering
     \begin{tabular}{@{}c@{}c@{}}
     \begin{subfigure}[b]{\errorplt\textwidth}
         \centering
         \includegraphics[width=\textwidth]{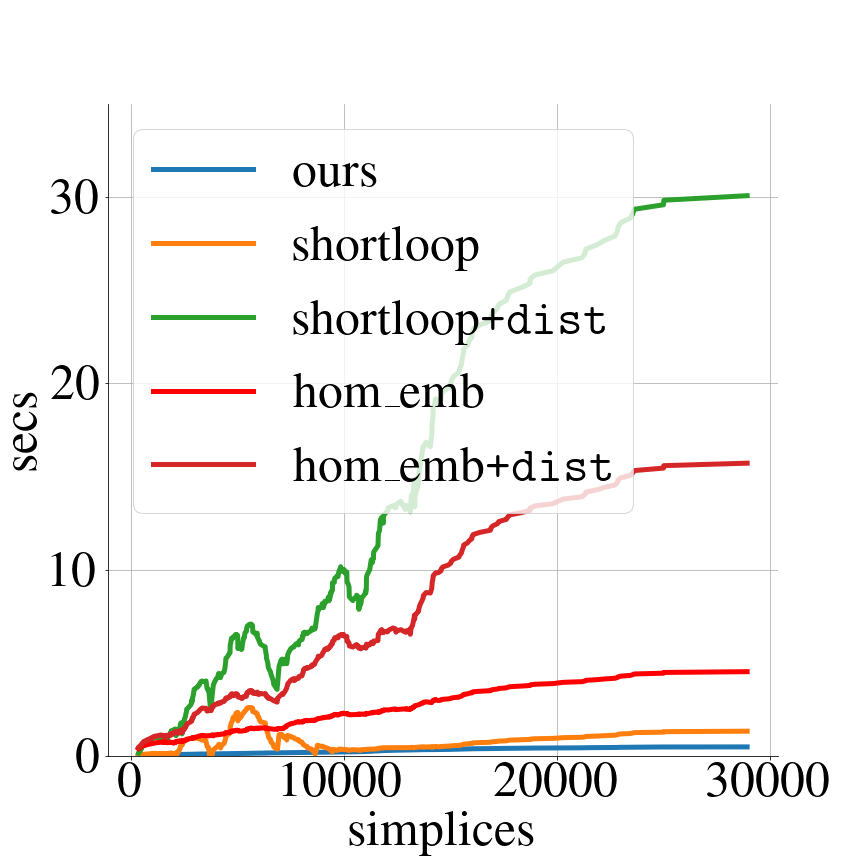}
         \label{fig:timeSimplices}
     \end{subfigure}
     \begin{subfigure}[b]{\errorplt\textwidth}
         \centering
         \includegraphics[width=\textwidth]{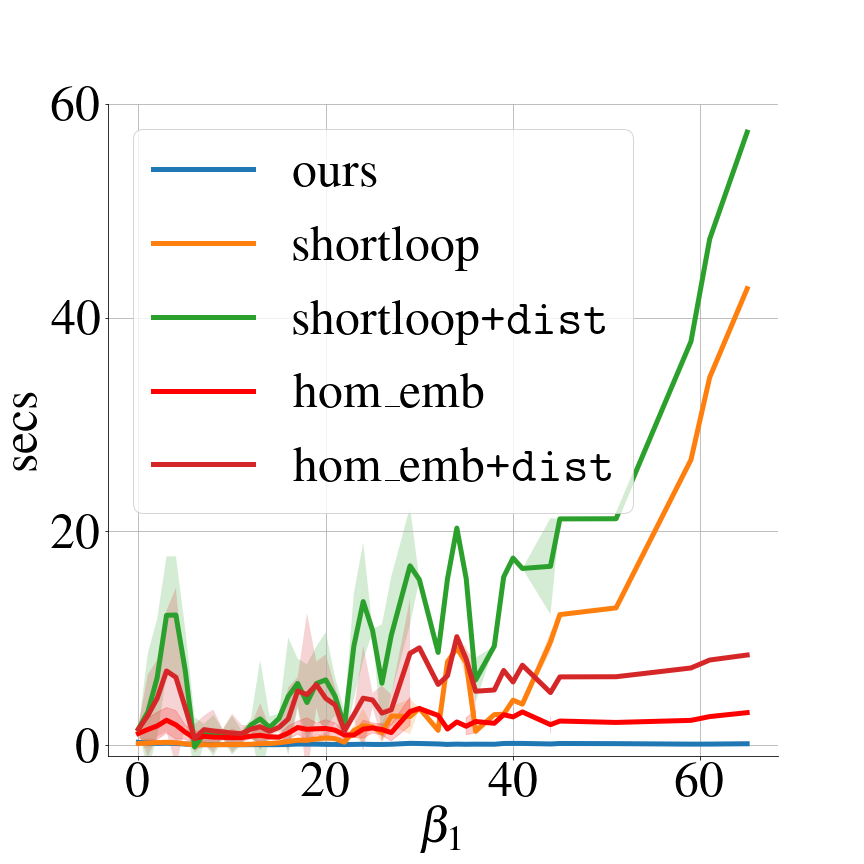}
         \label{fig:timeCycles}
     \end{subfigure}
     \end{tabular}
     \caption{Computation time as a function of the number of simplices~(left), and the number of generators~(right) for the 3D \texttt{TORI} dataset. We compare our trained model against the reference baseline, shortloop, and hom\_emb. \texttt{+dist} denotes additional post-processing required for obtaining distances.}
     \label{fig:timings}
\end{figure}

\subsection{Experimental Settings}\label{subsec:setting}

We used a GNN with 12 graph convolutional layers (for 2D as well as 3D), as described by Eq.~\eqref{eq:SFGC}, and 128 hidden units. We chose LeakyReLU activations ($\phi$ in Eq.~\eqref{eq:gnn}) with negative slope $r=0.02$ for the layers, and a hyperbolic tangent Tanh for the output. Neighbor activations are aggregated via a summation ($\bigoplus$ in Eq.~\eqref{eq:gnn}). Learnable weights undergo Kaiming uniform initialization~\cite{he2015delving}. Finally, node features are the result of concatenating the betti numbers describing the homology of the link at each simplex, with its 5-dimensional spectral embedding.

The dataset is split into training (80\%) and testing (20\%) sets and the models were trained for 1000 epochs, with a mini-batch size of 5 complexes using an Intel Xeon E5-2630 v.4 processor, a TITAN-X 64GB GPU and 64GB of RAM, using CUDA 10.1. The GNN model was implemented using the dgl library~\cite{wang2019dgl} with the Torch backend~\cite{paszke2017automatic}. All simplicial and homology computations were handled by the Gudhi library~\cite{gudhi:urm}. 

We apply a Laplacian smoothing post-processing step. Let $\bm{x}$ the output of the model, i.e. the inferred distances for each 1-simplex, and $\hat{L}=D^{-1/2}(D-A)D^{-1/2}$ the normalized graph Laplacian of the 1-skeleton of the complex $K$, i.e. the underlying graph spanned by the 0 and 1-simplices of $K$. The signal at the simplices are smoothed using 
$\bm{x}'=\bm{x}-\hat{L}\bm{x}$.

\subsection{Results}\label{subsec:results}

The main quantitative results can be found in Figure~\ref{fig:complplots} and Figure~\ref{fig:2D3Derrors} with qualitative examples in Figure~\ref{fig:complviz}. We report mean squared error (MSE) between the predicted and reference relative distances, with distances based on shortloop~\cite{dey2010approximating} acting as ground truth. We experimentally compare our method against a combinatorial baseline, hom\_emb~\cite{chen2021decomposition}. Since the range of $\hat{d}(\cdot)$ is within $[0,1]$, MSE can never exceed $1$, i.e. $100\%$ error. While additional baselines were considered, such as distr\_cover\_loc~\cite{Salehi2010Distributed}, unfortunately they do not provide code or empirical analysis that is easy to compare with. 

In 2D as well as 3D  our model learns the homology-parametrized distance function by achieving an MSE of $3.81\% \ (0.0381)$ (2D), and $4.47\% \ (0.0447)$ (3D), respectively, compared to $7.86\% \ (0.0786)$ (2D), and $8.89\% \ (0.0889)$ (3D), obtained by hom\_emb. Figure~\ref{fig:2D3Derrors} (first column) provides insights into the distribution of error at different distances (x-axis) from optimal generators. In 2D the error is monotonically increasing with the distance from the homology generators,  whereas in 3D the model performs slightly better at relative distances of about a third from the optimal cycles. Our method consistently outperforms hom\_emb at small and moderate distances ($<0.67$), whereas for the farthest distance range hom\_emb performs slightly better. In both settings, areas far away from the homology cycles attain the maximum mean MSE but never in excess of 15\%.

Figure~\ref{fig:2D3Derrors} also investigates the scalability of the model as the number of simplices, homology features ($\beta_1$) and maximum cycle lengths, increase (columns 2-4). The insets provide histograms of the respective parameter counts in the test sets to shed light into the standard deviation (shaded). The parameter values are non-uniformly represented in the test sets, partly explaining the larger variance towards the lower ends of the value ranges. Our model scales well in all three parameters and we observe that the error  decreases for larger numbers of simplices. The baseline, hom\_emb, overall scales similarly to our method (with the exception of scalability in terms of number of simplices), albeit exhibiting consistently higher MSE across all parameter scales. The maximum MSE of our method across all parameters never exceeded 12\%.  

Qualitative assessment of the model's performance is provided in Figure~\ref{fig:complviz} for examples from the test sets. The comparisons are arranged from lowest error (left) to maximum error (right) within our dataset.  The top row of figures visualize the predicted distances projected onto the complex while the bottom shows the ground truth. Cool areas indicate close proximity to an optimal homology generator. 

Figure~\ref{fig:complplots} plots reference distance values (orange)  and our model's output (blue) for each simplex against the rank of the simplex's distance from an optimal generator (X axis). Ideally, the blue curve should be monotonically increasing and should closely match the orange curve.

Both in 2D and 3D the model can handle multiple homology cycles of various lengths, even greater than the number of convolutional layers that usually dictate the receptive field of each simplex. Problematic appear to be the cases where components of the complexes have trivial homology, evident from the rightmost column in Figure~\ref{fig:complplots}. In such cases the model attempts to detect homological structure, where none exists.

Timing results are provided in Figure~\ref{fig:timings}, where computation time is presented as a function of the number of simplices, as well as the number of generators. Our trained model (blue) is more efficient than the two iterative baseline methods, shortloop (orange), and hom\_emb (red). We also exhibit time taken for other methods to post-process the results in order to obtain relative distances to the homology generators of interest (\texttt{+dist}). The cost of post-processing is independent of generators, but scales poorly with the number of simplices.

\subsection{Limitations and Conclusion}

Our model entifies simplices that are distant to homology cycles, aided by the shifted-inverted Hodge Laplacian based graph convolution and the simplified simplex adjacency construction. We experimented with using a Hasse graph analogue, as well as the original, non-shifted, non-inverted, Hodge Laplacians of the complex, but these models failed to learn. 

Another advantage of our model is that both large and small scale homology cycles are consistently localized. Although we restricted our analysis to $\mathcal{H}_1$ for ease of evaluation, the generality of our model allows direct extension to higher dimensional simplices.  

One drawback of our model is that it performs poorly when components of complexes  contain no higher dimensional homological information whatsoever. In such cases, it hallucinates homology cycles while they do not exist. 

Another limitation is the variance of the inferred function on the complex, as shown in Figure~\ref{fig:complplots}. This variance stems from two sources: First, the target function is piecewise constant; and second, the adjacency structure that we use to sparsify an otherwise complete graph (for creating the GNN computational graph) alters the spectrum of the kernel.

The choice of a spectral sparsification method with theoretical guarantees is an interesting avenue of future work. Needless to say, our results are promising even with a basic GNN architecture.


\section*{Acknowledgements}
VN is supported by EPSRC grant EP/R018472/1. Kartic Subr was supported by a Royal Society University Research Fellowship.

\bibliography{SFGNN.bib}

\end{document}